%% file: patch.tex
\documentclass[letterpaper, 10 pt, conference]{ieeeconf}  % Comment this line out if you need a4paper

\IEEEoverridecommandlockouts                              % This command is only needed if 
\pdfoutput=1
\usepackage{times}
\usepackage{graphicx}
\usepackage{epstopdf}
\usepackage{epsfig}
\usepackage{amsmath}
\usepackage{amssymb}
\usepackage{algorithmic}
\usepackage{algorithm}
\usepackage{multirow}
\usepackage{textcomp}
\usepackage{xspace}
\usepackage{color}
\usepackage[tight,scriptsize,sf,SF]{subfigure}
\usepackage[it,small]{caption}
\usepackage{array}
\usepackage{url}

\def \D {\mathbb{D}}

\def \0 {\mathbf{0}}
\def \1 {\mathbf{1}}

\makeatletter
\DeclareRobustCommand\onedot{\futurelet\@let@token\@onedot}
\def\@onedot{\ifx\@let@token.\else.\null\fi\xspace}

\def\eg{\emph{e.g}\onedot} 
\def\ie{\emph{i.e}\onedot} 
 
\def\etc{\emph{etc}\onedot} 
 
\def\etal{\emph{et al}\onedot} \def\st{s.t\onedot}
\newcolumntype{R}[1]{>{\raggedright\arraybackslash\hspace{0pt}}m{#1}}

\title{Watch-Bot: Unsupervised Learning for \\  Reminding Humans of Forgotten Actions}

\author{Chenxia Wu$^{1}$, Jiemi Zhang$^{2}$, Bart Selman$^{3}$, Silvio Savarese$^{4}$ and Ashutosh Saxena$^{5}$ % <-this % stops a space
\thanks{$^{1}$Chenxia Wu is with the Department of Computer Science, Cornell University and Stanford University.\texttt{\{chenxiawu\}@cs.cornell.edu}}
\thanks{$^{2}$Jiemi Zhang is with Didi research.\texttt{\{jmzhang10\}@gmail.com}}
\thanks{$^{3}$Bart Selman is with the Department of Computer Science, Cornell University.\texttt{\{selman\}@cs.cornell.edu}}
\thanks{$^{4}$Silvio Savarese is with the Department of Computer Science, Stanford University.\texttt{\{ssilvio\}@stanford.edu}}
\thanks{$^{5}$Ashutosh Saxena is with the Department of Computer Science, Cornell University and with Brain of Things, Inc.. \texttt{\{asaxena\}@cs.cornell.edu}}
}

\begin{document}

\maketitle
\thispagestyle{empty}
\pagestyle{empty}

\input{space_saver}
\input{abstract}

\input{introduction}

\input{relatedwork}
\input{overview}

\input{approach}

\input{learning}

\input{patching}

\input{experiments}

\input{conclusion}

{\small
\bibliographystyle{plain}
\bibliography{patch_short}
}

\end{document}

%% file: space_saver.tex
% Some illegal space-saving macros
% \parskip=1pt
  \abovedisplayskip 3.0pt plus2pt minus2pt%
 \belowdisplayskip \abovedisplayskip
\renewcommand{\baselinestretch}{0.98}

\newenvironment{packed_enum}{
\begin{enumerate}
  \setlength{\itemsep}{0pt}
  \setlength{\parskip}{0pt}
  \setlength{\parsep}{0pt}
}
{\end{enumerate}}

\newenvironment{packed_item}{
\begin{itemize}
  \setlength{\itemsep}{0pt}
  \setlength{\parskip}{0pt}
  \setlength{\parsep}{0pt}
}{\end{itemize}}

\newlength\savedwidth
\newcommand\whline[1]{\noalign{\global\savedwidth\arrayrulewidth
                               \global\arrayrulewidth #1} %
                      \hline
                      \noalign{\global\arrayrulewidth\savedwidth}}
\renewcommand\multirowsetup{\centering}

\newlength{\sectionReduceTop}
\newlength{\sectionReduceBot}
\newlength{\subsectionReduceTop}
\newlength{\subsectionReduceBot}
\newlength{\abstractReduceTop}
\newlength{\abstractReduceBot}
\newlength{\captionReduceTop}
\newlength{\captionReduceBot}
%\newlength{\nameReduceTop}
\newlength{\subsubsectionReduceTop}
\newlength{\subsubsectionReduceBot}
\newlength{\headerReduceTop}
% Negative space for figures set at the bottom of a block of figs
\newlength{\figureReduceBot}

\newlength{\horSkip}
\newlength{\verSkip}

\newlength{\equationReduceTop}

\newlength{\figureHeight}
\setlength{\figureHeight}{1.7in}

%\newlength{\figureFraction}
\setlength{\horSkip}{-.09in}
\setlength{\verSkip}{-.1in}
%\setlength{\figureFraction}{.195}

% figureReduceBot is for figures which are set above text, since latex
% likes putting a lot of space under those
\setlength{\figureReduceBot}{-0.15in}
\setlength{\headerReduceTop}{0in}
\setlength{\subsectionReduceTop}{-0.02in}
\setlength{\subsectionReduceBot}{-0.02in}
\setlength{\sectionReduceTop}{-0.02in}
\setlength{\sectionReduceBot}{-0.01in}
\setlength{\subsubsectionReduceTop}{-0.06in}
\setlength{\subsubsectionReduceBot}{-0.05in}
\setlength{\abstractReduceTop}{-0.05in}
\setlength{\abstractReduceBot}{-0.10in}

\setlength{\equationReduceTop}{-0.1in}

\setlength{\captionReduceTop}{-0.06in}
\setlength{\captionReduceBot}{-0.07in}

%% file: abstract.tex
% !TEX root = patch.tex

\begin{abstract}

We present a robotic system that watches a human using a Kinect v2 RGB-D sensor, detects what he forgot to do while performing an activity, and if necessary reminds the person using a laser pointer to point out the related object. Our simple setup can be easily deployed on any assistive robot.

Our approach is based on a learning algorithm trained in a purely unsupervised setting, which does not require any human annotations. This makes our approach scalable and applicable to variant scenarios. Our model learns the action/object co-occurrence and action temporal relations in the activity, and uses the learned rich relationships to infer the forgotten action and the related object. We show that our approach not only improves the unsupervised action segmentation and action cluster assignment performance, but also effectively detects the forgotten actions on a challenging human activity RGB-D video dataset. In robotic experiments, we show that our robot is able to remind people of forgotten actions successfully.

\end{abstract}

%% file: introduction.tex
% !TEX root = patch.tex

\section{Introduction}\label{sec:intro}

\begin{figure}[t]
  \begin{center}
  \includegraphics[width=0.85\linewidth]{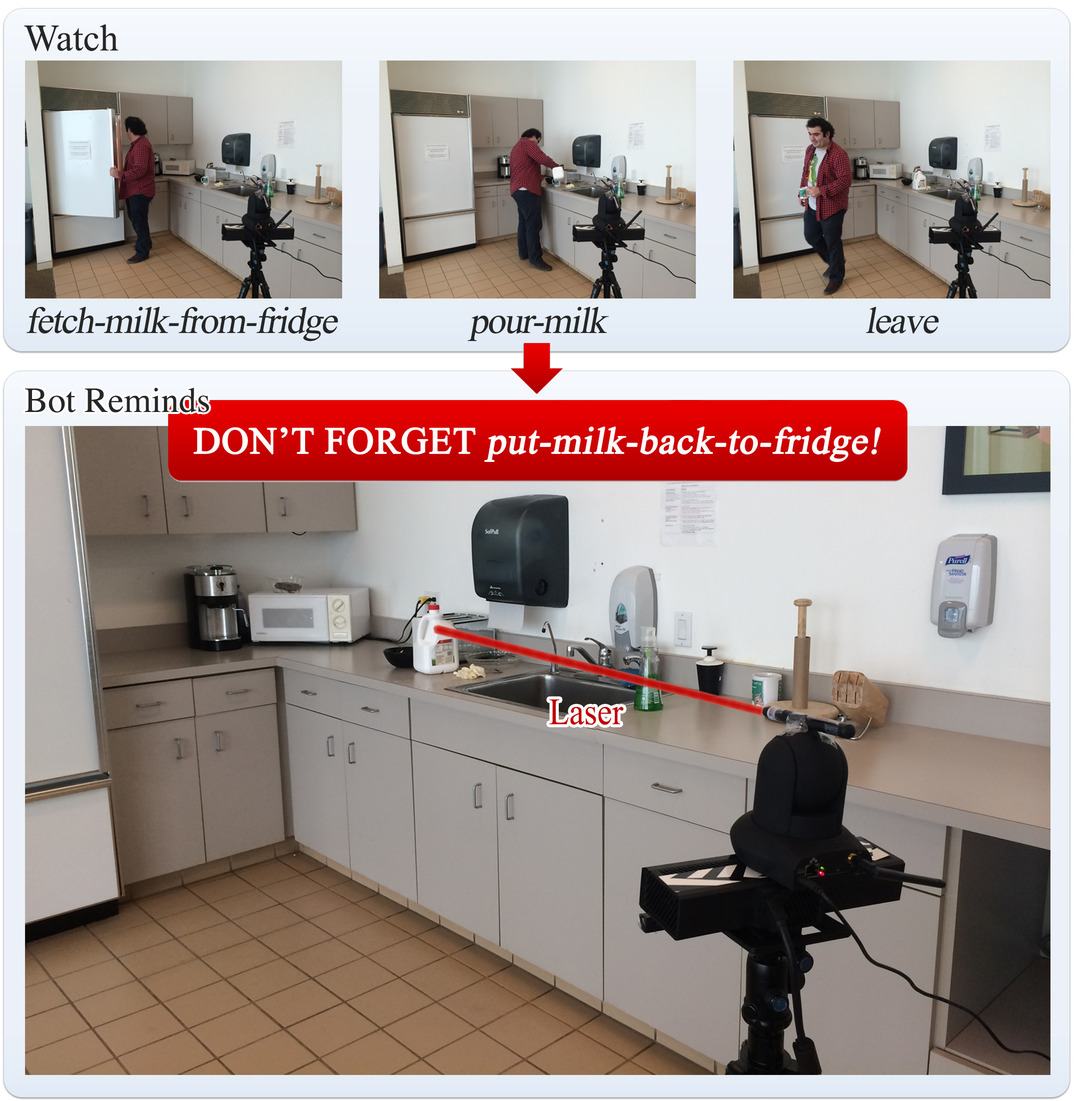}

 \caption{Our Watch-Bot watches what a human is currently doing, and uses our unsupervised learning model to detect the human's forgotten actions. Once a forgotten action detected (\emph{put-milk-back-to-fridge} in the example), it points out the related object (\emph{milk} in the example) by the laser spot in the current scene.}
 \label{fig:tr}
 \end{center}
 \vspace{-0.35in}
\end{figure}

The average adult forgets three key facts, chores or events every day~\cite{forget}. Hence it is important for a personal robot to be able to detect not only what a human is currently doing but also what he forgot to do. For example in Fig.~\ref{fig:tr}, someone fetches milk from the fridge, pours the milk to the cup, takes the cup and leaves without putting back the milk, then the milk would go bad. In this paper, we focus on detecting these forgotten actions in the complex human activities for a robot, which learns from a completely unlabeled set of RGB-D videos.
%Many previous works focus on improving the former ability of action/activity recognition~\cite{Aggarwal_2011_SURVEY,Sung_2012_ICRA,Koppula_2013_IJRR,Piyathilaka_2015_FSR}.

%The challenge in this paper is two-fold: Can a robot remind a human of forgotten actions in a simple but effective way, and can the robot do so by learning from a completely unlabeled set of RGB-D videos?

There are a large number of works on vision-based human activity recognition for robots. These works infer the semantic label of the overall activity or localize actions in the complex activity for better human-robot interactions~\cite{Losch_2007_RO,Agarwal_2012_RSS,Chrungoo_2014_SR}, assistive robotics~\cite{Jiang_2014_RSS,Yang_2015_AAAI}. Given the input RGB/RGB-D videos~\cite{Sung_2012_ICRA,Koppula_2013_IJRR,Chen_2014_ICRA}, or 3D human joint motions~\cite{Mansur_2011_ICRA,Piyathilaka_2015_FSR},  or from other inertial/location sensors~\cite{Chen_2012_SMC,Min_2011_SMC}, they train the perception model using fully or weekly labeled actions~\cite{Koppula_2013_IJRR,Bojanowski_2014_ECCV,Hu_2014_RSS}, or locations of annotated human/their interactive objects~\cite{Tian_2013_CVPR,Ni_2014_CVPR}. Recently, there are some other works on anticipating human activities for reactive robotic response~\cite{Koppula_2013_RSS,Jiang_2014_RSS}. However, to enable a robot to remind people of forgotten things, it is challenging to directly use these approaches especially in a completely unsupervised setting.

%And they model different types of information, such as the temporal relations between actions~\cite{Shi_2011_IJCV,Pirsiavash_2014_CVPR,Wang_2014_CVPR}, the human and the interactive object appearances and relations~\cite{Wu_2014_CVPR,Koppula_2013_ICML,Wang_2014_CVPR}.

\begin{figure*}[t]
  \begin{center}
  \subfigure[Robot System.]{
  \includegraphics[height=2.3in]{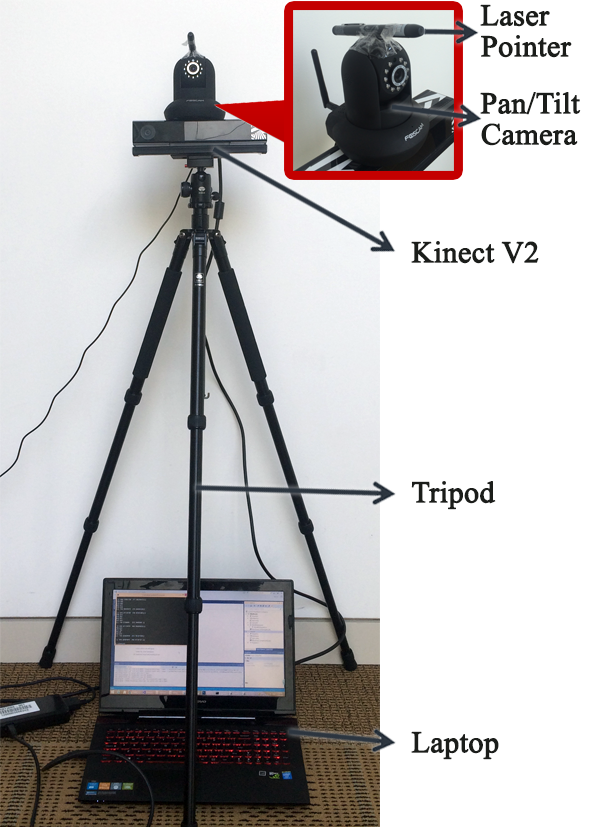}
  }
  \subfigure[System Pipeline.]{
  \includegraphics[height=2.3in]{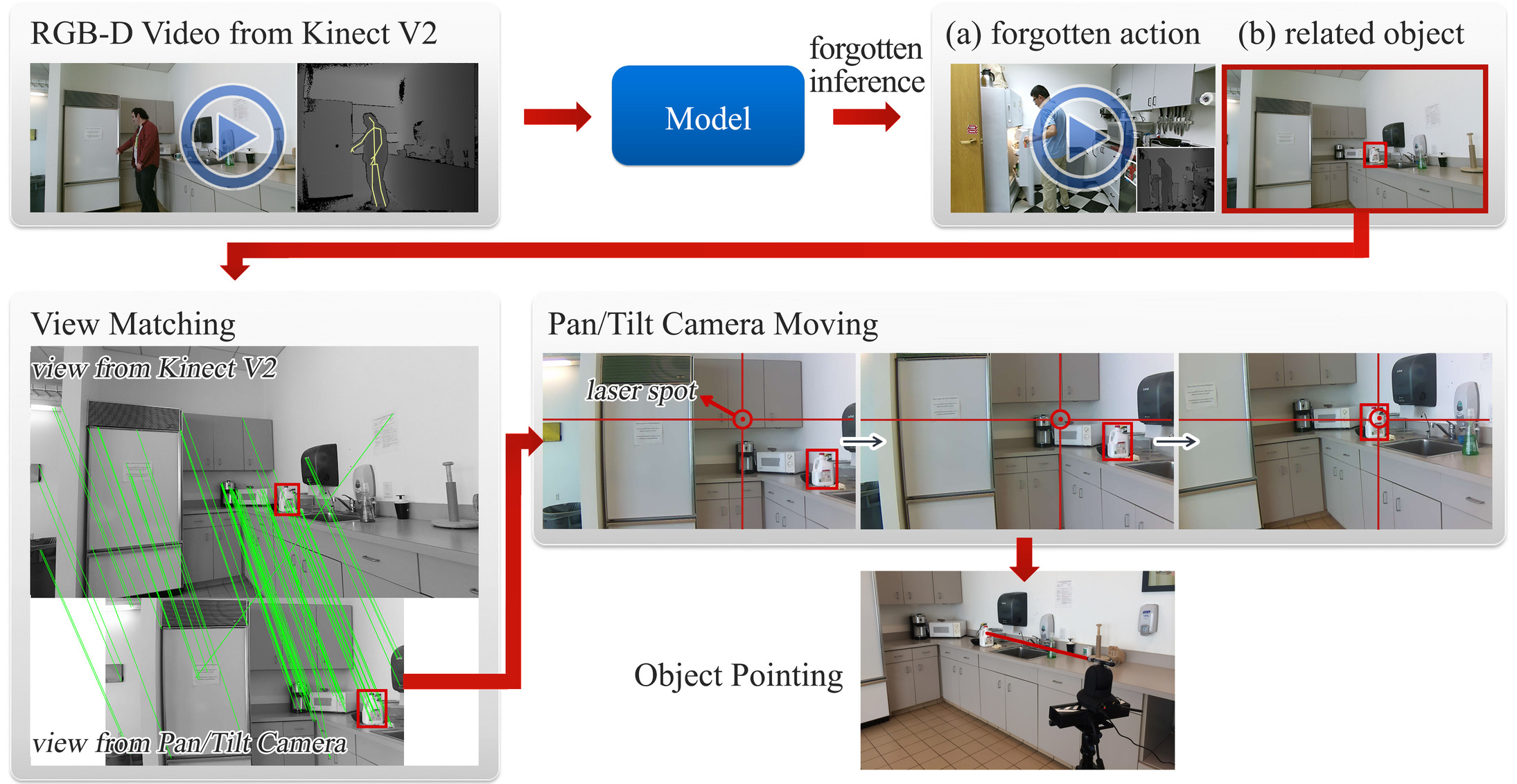}
  }
\vspace{-0.1in}
 \caption{ (a). Our Watch-Bot system. It consists of a Kinect v2 sensor that inputs \mbox{RGB-D} frames of human actions, a laptop that infers the forgotten action and the related object, a pan/tilt camera that localizes the object, mounted with a fixed laser pointer that points out the object. (b). The system pipeline.  The robot first uses the learned model to infer the forgotten action and the related object based on the Kinect's input. Then it maps the view from the Kinect to the pan/tilt camera so that the bounding box of the object is mapped in the camera's view. Finally, the camera pan/tilt until the laser spot lies in the bounding box of the target object.}
 \label{fig:system}
   \vspace{-0.3in}
 \end{center}
\end{figure*}

Our goal is to enable a robot, that we call Watch-Bot, to detect humans' forgotten actions as well as localize the related object in the current scene. The robot consists of a Kinect v2 sensor, a pan/tilt camera (which we call camera for brevity in this paper) mounted with a laser pointer, and a laptop (see Fig.~\ref{fig:system}(a)). This setup can be easily deployed on any assistive robot. Taking the example in Fig.~\ref{fig:tr}, if our robot sees a person fetch a milk from the fridge, pour the milk, and leave without putting the milk back to the fridge. Our robot would first detect the forgotten action and the related object (the milk), given the input \mbox{RGB-D} frames and human skeletons from the Kinect; then map the object from the Kinect's view to the camera's view; finally pan/tilt the camera till its mounted laser pointer pointing to the milk.

In real robotic applications, people perform a very wide variety of actions. These are hard to learn from existing videos on the Internet and there are few with annotations of actions or objects.  So we propose a probabilistic learning model in a completely unsupervised setting, which can learn actions and relations directly from the data without any annotations, only given the input \mbox{RGB-D} frames with tracked skeletons from Kinect v2 sensor. %So it is scalable and easy to be used in different scenarios. 

We model an activity video as a sequence of actions, so that we can understand which actions have been taken, \eg, the example activity contains four actions: \emph{fetch-milk-from-fridge}, \emph{pour}, \emph{put-milk-back-to-fridge}, and \emph{leave}.\footnote{In the training, we do not know these action semantic labels. Instead we assign the action cluster index.} For detecting the forgotten action and reminding, we model the co-occurrence between actions and the interactive objects, as well as the temporal relations between these segmented actions, \eg, action \emph{fetch-milk-from-fridge} often co-occurs with and is temporally after action \emph{put-milk-back-to-fridge}, and object \emph{milk} occurs in both actions. Using the learned actions and relations, we infer the forgotten actions and localize the related objects, \eg, \emph{put-milk-back-to-fridge} might be forgotten as previously seen \emph{fetch-milk-from-fridge} before \emph{pouring}, and seen \emph{leaving} indicates he really forgot to do, also \emph{milk} is the object interacted in the forgotten action.

We evaluate our approach extensively on a large \mbox{RGB-D} human activity dataset recorded by Kinect v2~\cite{Wu_2015_CVPR}. The dataset contains $458$ videos of human daily activities as compositions of multiple actions interacted with different objects, in which people forgot actions in $222$ videos. %The activities are performed by different subjects in different environments with complex backgrounds.
We show that our approach not only improves the action segmentation and action cluster assignment performance, but also obtains promising results of forgotten action detection. Moreover, we show that our Watch-Bot is able to remind humans of forgotten actions in the real-world robotic experiments.

\if
We consider modeling human activities containing a sequence of actions (see an example in Fig.~\ref{fig:tr}),
% mprise a complex sequence of action
 as perceived by an \mbox{RGB-D} sensor in home and office environments.
 % The type of activities often comprise a complex sequence of actions and
 % include interactions with different objects.
 In the complex human activity such as \emph{warming milk} in the example, there are not only short-range action relations, \eg, \emph{microwaving} is often followed by \emph{fetch-bowl-from-oven}, but there are also long-range action relations, \eg,  \emph{fetch-milk-from-fridge} is strongly related to \emph{put-milk-back-to-fridge} even though several actions occur between them.

The challenge that we undertake in this paper is:  Can an algorithm learn about the aforementioned relations in the activities when just given a completely \emph{unlabeled} set of \mbox{RGB-D} videos?

Most previous works focus on action detection in a supervised learning setting. In the training, they are given fully labeled actions in videos~\cite{Liu_2011_CVPR,Sadanand_2012_CVPR,Schiele_2012_CVPR},  or weakly supervised action labels~\cite{Duchenne_2009_ICCV,Bojanowski_2014_ECCV}, or locations of human/their interactive objects~\cite{Laptev_2007_ICCV,Tian_2013_CVPR,Ni_2014_CVPR}. Among them, the temporal structure of actions is often discovered by Markov models such as Hidden Markov Model (HMM)~\cite{Tang_2012_CVPR} and semi-Markov~\cite{Hoai_2011_CVPR, Shi_2011_IJCV}, or by linear dynamical systems~\cite{Bhattacharya_2014_CVPR}, or by hierarchical grammars~\cite{Pirsiavash_2014_CVPR,Vo_2014_CVPR,Kuehne_2014_CVPR,Wang_2014_CVPR,Assari_2014_CVPR}, or by other spatio-temporal representations~\cite{Ke_2007_ECCV,Niebles_2010_ECCV,Klaser_2010_ECCV,Koppula_2013_ICML}. Most of these works are based on RGB features and only model the short-range relations between actions (see section~\ref{sec:re} for details). %Some other works focus on discovering mid-level discriminative patches~\cite{Jain_2013_CVPR} or local motion primitives~\cite{Yang_2013_TPAMI} of actions.

%Moreover, many hierarchical models  parse a video hierarchically into multiple action instances and sub-actions.

Different from these approaches, we consider a completely unsupervised setting. The novelty of our approach is the ability to model the long-range action relations in the temporal sequence,
% \eg, \emph{fetch-book} is strongly related to \emph{put-back-book} with a long \emph{reading} between them.
% We model not only the mapping of human skeleton and \mbox{RGB-D} features to actions,
by considering pairwise action co-occurrence and temporal relations, \eg, \emph{put-milk-back-to-fridge} often co-occurs with and temporally after \emph{fetch-milk-from-fridge}. We also use the more informative human skeleton and \mbox{RGB-D} features, which show  higher performance
%  been shown the advantages
over just RGB features for action recognition~\cite{Koppula_2013_RSS,Wu_2014_CVPR,Lin_2014_CVPR}.

%Some other works have focused on segmenting on low-level visual features \cite{ZZZ}, and also on learning mid-level discriminative video patches \cite{gupta}. Markov Models (e.g., HMM) are often used to model the action transitions in the activities.  \todo{More}Different from previous linear models on action recognition~\cite{Tang_2012_CVPR, Hoai_2011_CVPR, Shi_2011_IJCV,Bhattacharya_2014_CVPR}, which depend on local relations between adjacent clips or actions, and grammar-based models~\cite{Pirsiavash_2014_CVPR,Vo_2014_CVPR,Kuehne_2014_CVPR}, which parse a video into multiple action instances and sub-actions using hierarchical relations.

In order to capture the rich structure in the activity, we draw strong parallels with the work done on document modeling from natural language (e.g., ~\cite{Blei_2003_LDA}). We consider an activity video as a document,  which consists of a sequence of short-term action clips as \emph{action-words}. And an activity is about a set of \emph{action-topics} indicating which actions are present in the video, such as \emph{fetch-milk-from-fridge} in the \emph{warming milk} activity. Action-words are drawn from these action-topics and has a distribution for each topic. Then we model the following (see Fig.~\ref{fig:pipeline}):
\vspace{-0.06in}
\begin{packed_item}
%\item \emph{Action-topics.}  %We decompose an action video to continuous overlapping fixed-length temporal clips. The visually similar clips are mapped to a single word---drawing an analogy with the words in natural language text.  These action-words belong to a dictionary that we learn.

%Action-words are drawn from these action-topics and has a distribution for each topic. %\eg, \emph{fetch-milk-from-fridge} might contain action-words of \emph{walking}, \emph{open-fridge}, \emph{put-in-milk}, \emph{close-fridge}, \etc.
% \item \emph{Action-topics.}  An activity video is a document, and it is about a set of topics, \eg, \emph{fetch-milk-from-fridge}, \emph{microwaving} in the \emph{warming milk} activity.  Action-words are drawn from these action-topics and has a distribution for each topic.
\item \emph{Action co-occurrence.}  Some actions often co-occur in the same activity. We model the co-occurrence by adding correlated topic priors to the occurrence of action-topics, \eg, \emph{fetch-milk-from-fridge} and \emph{put-milk-back-to-fridge} has strong correlations.
%We draw the occurrence of action-topics from correlated topic priors. The correlations capture the co-occurrence of actions, \eg, \emph{fetch-milk-from-fridge} and \emph{put-milk-back-to-fridge} has strong correlations.
% There are certain correlations that we want to model
\item \emph{Action temporal relations.}   Some actions often causally follow each other, and actions change over time during the activity execution. We model the relative time distributions between every action-topic pair to capture the temporal relations.

% Action-topics have a strong temporal structure, similar to words having structure in a document to form sentences.
\end{packed_item}
\vspace{-0.06in}
We first show that our model is able to learn meaningful representations from the unlabeled activity videos. We use the model to segment videos to segments assigned with action-topics. We show that the learned action-topics are semantic meaningful by mapping them to ground-truth semantic action classes and evaluating the action labeling performance.

We then also show that our model can be used to detect forgotten actions in the activity, a new application that we call \emph{action patching}. We show that the learned co-occurrence and temporal relations are very helpful to infer the forgotten actions by evaluating the patching accuracy.
\fi

%% file: relatedwork.tex
% !TEX root = patch.tex

\section{Related Work}\label{sec:re}
%There is a large number of works on action recognition, which can be referred in recent surveys~\cite{Aggarwal_2011_SURVEY}. %In this section, we cover the most related approaches.

Most previous works focused on recognizing human actions for both robotics~\cite{Losch_2007_RO,Koppula_2013_IJRR,Chen_2014_ICRA} and computer vision~\cite{Ke_2007_ECCV,Tang_2012_CVPR,Aggarwal_2011_SURVEY}. They model different types of information, such as the temporal relations between actions~\cite{Pirsiavash_2014_CVPR,Wang_2014_CVPR}, the human and the interactive object appearances and relations~\cite{Koppula_2013_ICML,Wang_2014_CVPR}. Yang~\etal.~\cite{Yang_2015_AAAI} presented a system that learns manipulation action plans for robot from unconstrained youtube videos. Hu~\etal.~\cite{Hu_2014_RSS} proposed an activity recognition system trained from soft labeled data for the assistant robot. Chrungoo~\etal.~ \cite{Chrungoo_2014_SR} introduced a human-like stylized gestures for better human-robot interaction.
% ,and contrast them from conventional human daily activities. 
Piyathilaka~\etal.~\cite{Piyathilaka_2015_FSR} used 3D skeleton features and trained dynamic bayesian networks 
% to human activities 
for domestic service robots. However, it is challenging to directly use these approaches for inferring the forgotten actions.

Recently, there are works on anticipating human activities and they performed well for assistant robots~\cite{Koppula_2013_RSS,Jiang_2014_RSS}. They modeled the object affordances and object/human trajectories to discriminate different actions in past activities and anticipate future actions. However, in order to detect forgotten actions, we also need to consider actions after it such as \emph{boiling water} indicates \emph{filling kettle} before it.
%we need to consider not only past-to-future relations used in anticipation, but also future-to-past relations such as \emph{boiling water} indicates \emph{filling kettle} before it.

%Our work is also related to human pose estimation or human parts recognition based robots. Agarwal~\etal.~\cite{Agarwal_2012_RSS} estimated human dynamics by inertial and kinematic parameter estimation using an uncalibrated monocular video camera. Gioioso~\etal.~\cite{Gioioso_2012_RSS} presented a method to map human synergies onto robotic hands by using a virtual object.

The output laser spot on object is also related to the work `a clickable world'~\cite{Nguyen_2008_RSJ}, which selects the appropriate behavior to execute for an assistive
object-fetching robot using the 3D location of the click by the laser pointer. Differently, we keep the laser pointer fixed on top of the camera, and pan/tilt the camera iteratively to point out the target object using a real-time view matching.

Most of these works rely on supervised learning given fully labeled actions, or weakly supervised action labels, or locations of human/their interactive objects. Differently, our robot uses a completely unsupervised learning setting that trains model only on Kinect's output \mbox{RGB-D} videos. Our model is based on our previous work~\cite{Wu_2015_CVPR}, which presents a Casual Topic Model to model action relations in the complex activity. In this paper, we further introduce the human interactive object and its relations to actions, so that the robot can localize the related object. We then design a robotic system using the model to kindly remind people.

\if

Most previous works are focused on recognizing human actions. Many of them are supervised~\cite{Laptev_2007_ICCV,Duchenne_2009_ICCV,Niebles_2010_ECCV,Liu_2011_CVPR,Sadanand_2012_CVPR,Tian_2013_CVPR,Bojanowski_2014_ECCV}. Among them, the linear models~\cite{Tang_2012_CVPR,Hoai_2011_CVPR, Shi_2011_IJCV,Bhattacharya_2014_CVPR} are the most popular, which focus on modeling the action transitions in the activities. More complex hierarchical relations~\cite{Pirsiavash_2014_CVPR,Vo_2014_CVPR,Kuehne_2014_CVPR,Wang_2014_CVPR} or graph relations~\cite{Assari_2014_CVPR} are considered in modeling actions in the complex activity. Although they have performed well in recognizing human actions, most of them rely on local relations between adjacent clips or actions that ignore the long-term action relations. However, these long-term action relations are also important to detecting forgotten actions, such as \emph{fetch-book} strongly indicates \emph{put-back-book} even though there is a long \emph{reading} between them.

%There are also some works focusing on detecting local action patches, primitives or trajectories~\cite{Jain_2013_CVPR,Yang_2013_TPAMI,Narayan_2014_CVPR} without considering the high-level action relations.

%of actions with a large time interval can also be related. %, \eg, \emph{fetch-book} is strongly related to \emph{put-back-book} with a long \emph{reading} between them.
%Recently, more complex hierarchical relations are considered in many hierarchical models~\cite{Pirsiavash_2014_CVPR,Vo_2014_CVPR,Kuehne_2014_CVPR,Wang_2014_CVPR}. %However, actions do not necessarily form a hierarchy but a graph.
%Moreover, Assari \etal~\cite{Assari_2014_CVPR} presents a richer representation of action videos based on the co-occurrence of action concepts using a graph. %while temporal information is not well captured.

There also exist some unsupervised approaches on action recognition. Yang \etal~\cite{Yang_2013_TPAMI} develop a meaningful representation by discovering local motion primitives in an unsupervised way, then a HMM is learned over these primitives. Jones \etal~\cite{Jones_2014_CVPR} propose an unsupervised dual assignment clustering on the dataset recorded from two views.

Different from these approaches, we use the richer human skeleton and \mbox{RGB-D} features rather than the RGB action features~\cite{Wang_2011_CVPR,Kantorov_2014_CVPR}. We model the pairwise action co-occurrence and temporal relations in the whole video, thus relations are considered globally and completely with the uncertainty.  These learned relations can be used to infer the forgotten actions without any manual annotations.

Action recognition using human skeletons and \mbox{RGB-D} camera have shown the advantages over RGB videos in many works. Skeleton-based approach focus on proposing good skeletal representations~\cite{Schiele_2012_CVPR, Vemulapalli_2014_CVPR,Wu_2014_CVPR,Lin_2014_CVPR}. Besides of the human skeletons, we also detect the human interactive objects in an unsupervised way to provide more discriminate features. Object-in-use contextual information has been commonly used for recognizing actions~\cite{Koppula_2013_RSS,Koppula_2013_ICML,Ni_2014_CVPR,Wang_2014_CVPR}. Most of them depend on correct object tracking or local motion changes. They lost the high-level action relations which can be captured in our model and is important to action patching.

Our work is also related to the topic models. LDA~\cite{Blei_2003_LDA} was the first hierarchical Bayesian topic model and widely used in different applications. The correlated topic models~\cite{Blei_2007_CTM,Kim_2011_NIPS} add the priors over topics to capture topic correlations. The topic model considering the topic specific distributions of the absolute time of word capture how the topic changes over time~\cite{Wang_2006_KDD} and has been applied to action recognition~\cite{Tanveer_2009_BMVC}. Differently, our model considers both correlations and the relative time distributions between topics rather than the absolute time, which captures richer information of action structures for patching in the complex human activity.

\fi

%Action temporal segmentation and recognition is often done with linear models such as hidden markov model (HMM)~\cite{Tang_2012_CVPR}, semi-Markov~\cite{Hoai_2011_CVPR, Shi_2011_IJCV}. They only consider the relations between adjacent frames or segments, while some actions with a large time gap might be correlated, \eg, \emph{fetch-book} and \emph{put-back-book} are strongly correlated while there is a long \emph{reading} between them.  Recently, inspired by natural language processing, grammars are used to hierarchically parse a video into multiple action instances and sub-actions~\cite{Pirsiavash_2014_CVPR,Vo_2014_CVPR,Kuehne_2014_CVPR} . However, they all require fully-labeled actions in training stage, which is more expensive than our unsupervised setting. Our model can capture semantic and temporal relations in the whole video by a cheap unsupervised learning. 

%% file: overview.tex
% !TEX root = patch.tex

\section{Watch-Bot System}\label{sec:ov}

We outline our Watch-Bot system in this section (see Fig.~\ref{fig:system}). Our goal is to detect what people forgot to do given the observation of his poses and interacted objects. The robot consists of a Kinect v2 sensor, a pan/tilt camera mounted with a laser pointer, and a laptop. The input to our system is \mbox{RGB-D} human activity videos with the tracked 3D joints of human skeletons from Kinect v2. Then we use an unsupervised trained learning model (see Section~\ref{sec:apc}) to infer the forgotten action and localize the related object in the Kinect's view. After that, we map the object bounding box from the Kinect's view to the camera's view. %Since the laser pointer is fixed on top of the camera, the laser spot is at a fixed position in the camera's view. So 
Finally, we pan/tilt the camera until the laser spot lies within the target object in its view (see Section~\ref{sec:ap}).

%Once detecting the forgotten action, the laptop displays a video clip of the same type of action retrieved from the learning database, as well as the frame of the current scene containing the bounding box of the related object. Finally, the pan/tilt camera moves till its bound laser pointer pointing to the related object in the current scene by mapping its view to Kinect's view.

\begin{figure}[t]
  \begin{center}
  \includegraphics[width=0.8\linewidth]{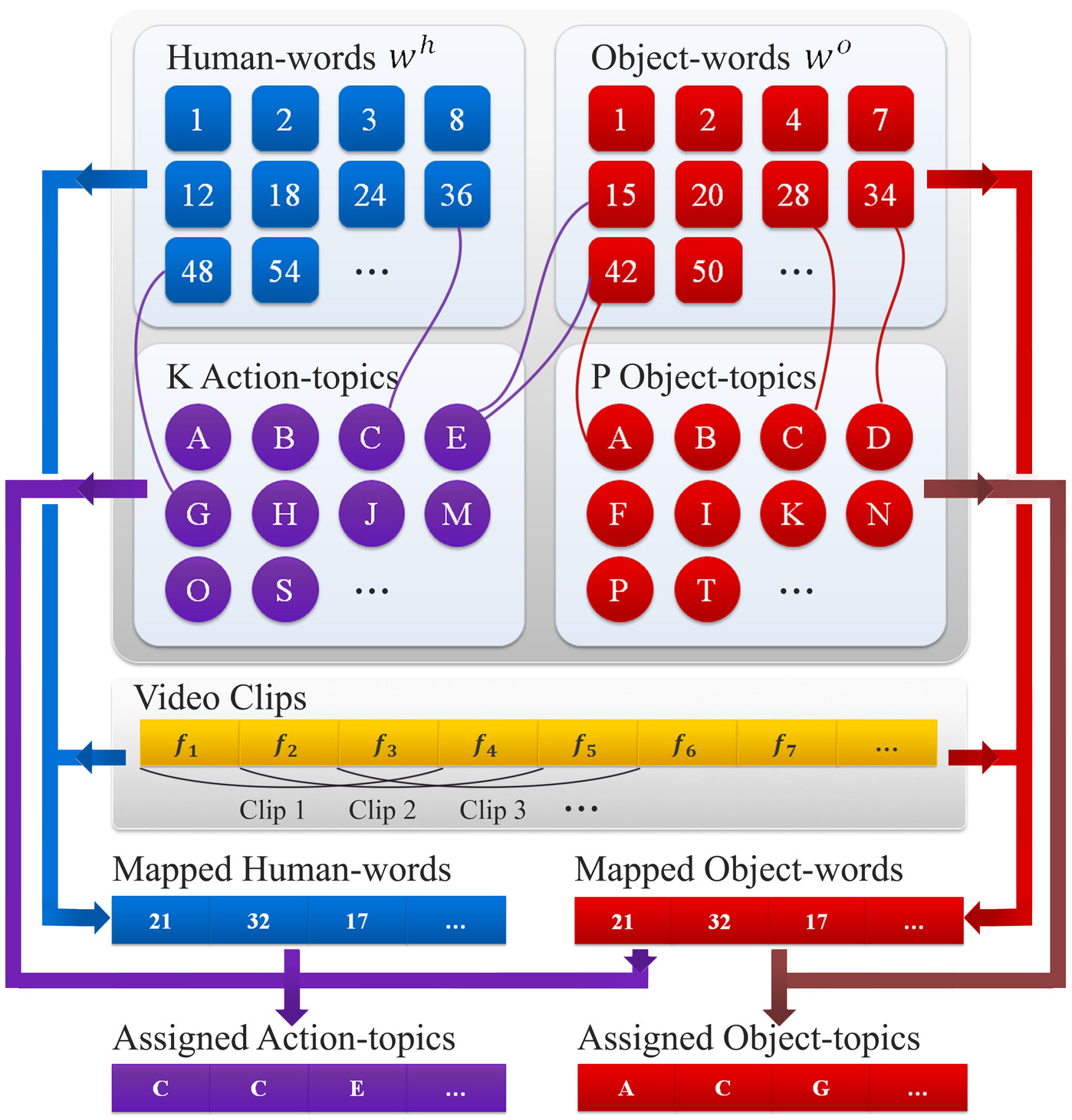}

 \caption{Video representation in our approach. A video is first decomposed into a sequence of overlapping fixed-length temporal clips.  The human-skeleton-trajectories/interactive-object-trajectories from all the clips are clustered to form the human-dictionary/object-dictionary. Then the video is represented as a sequence of human-word and object-word indices by mapping its human-skeleton-trajectories/interactive-object-trajectories to the nearest human-words/object-words in the dictionary. Also, an activity video is about a set of action-topics/object-topics indicating which actions are present and which object types are interacted.} \label{fig:vrep}
 \end{center}
 \vspace{-0.3in}
\end{figure}

{\bf Video Representation.} To detect the action structure in the complex activity video, we propose a video representation that draws parallels to document modeling in the natural language~\cite{Blei_2003_LDA} (illustrated in Fig.~\ref{fig:vrep}). We first decompose a video into a sequence of overlapping fixed-length temporal clips. We then extract the human-skeleton-trajectory features and the interactive-object-trajectory features from the clips. In order to build a compact representation of the activity video, we represent it as a sequence of words. We use $k$-means to cluster the human-skeleton-trajectories/interactive-object-trajectories from all the clips to form a \emph{human-dictionary} and an \emph{object-dictionary}, where we use the cluster centers as \emph{human-words} and \emph{object-words}. Then, the video can be represented as a sequence of human-word and object-word indices by mapping its human-skeleton-trajectories/interactive-object-trajectories to the nearest human-words/object-words in the dictionary. Also, an activity video is about a set of \emph{action-topics} indicating which actions are present in the video, and a set of \emph{object-topics} indicating which object types are interacted.

{\bf Visual Features.} We extract both human-skeleton-trajectory features and the interactive-object-trajectory features from the output by the Kinect v2. The new Kinect v2 has high resolution of RGB-D frames (RGB: $1920 \times 1080$, depth: $512 \times 424$) and improved body tracking of $25$ body joints of human skeletons.%\footnote{\url{http://www.microsoft.com/en-us/kinectforwindows/develop/}} . 

\if 0
Let $X_c=\{x_1^c,x_2^c,\cdots,x_{25}^c\}$ be the 3D coordinates of $25$ joints of a skeleton in the current frame.  We first compute the cosine of the angles between the connected parts in each frame: $\alpha_i=p_{i+1}\cdot p_i/|p_{i+1}|\cdot|p_i|$, where $p_i=x_{i+1}-x_i$ is the body part. The change of the joint coordinates and angles can well capture the human body movements. So we extract the motion features and off-set features~\cite{Wu_2014_CVPR} by computing their Euclidean distances $\D(,)$ to previous frame $f^x_{c,c-1}, f^\alpha_{c,c-1}$ and the first frame $f^x_{c,1},f^\alpha_{c,1}$ in the clip:
 \begin{equation}
 \begin{split}
 &f^x_{c,c-1}=\{\D(x_i^c,x_i^{c-1})\}_{i=1}^{25},\ f^\alpha_{c,c-1}=\{\D(\alpha_i^c,\alpha_i^{c-1})\}_{i=1}^{25};\\
&f^x_{c,1}=\{\D(x_i^c,x_i^{1})\}_{i=1}^{25},\ f^\alpha_{c,1}=\{\D(\alpha_i^c,\alpha_i^{1})\}_{i=1}^{25}.\notag
 \end{split}
 \end{equation}
 Then we concatenate all $f^x_{c,c-1},f^\alpha_{c,c-1},f^x_{c,1},f^\alpha_{c,1}$ as the human-skeleton-trajectory features of the clip.
\fi

We first extract the human-skeleton-trajectory features of the clip as in~\cite{Wu_2015_CVPR}. Then we extract the human interactive-object-trajectory based on the human hands, image segmentation, motion detection and tracking. %The extracted interactive-object-trajectory can help discriminate the different human actions with similar body motions such as \emph{fetch-book} and \emph{turn-on-monitor}, and localize the related object in patching.
We collect the bounding boxes enclosing the potential interested objects from super-pixels output by a fast edge detection approach~\cite{Dollar_2013_ICCV} on both RGB and depth images. We apply the moving foreground mask~\cite{Chris_CVPR_1999} to remove the unnecessary steady backgrounds and select those segments within a distance to the human hand joints in both 3D points and 2D pixels.

We then track the bounding box in the clip using SIFT matching and RANSAC to get the trajectories. We use the closest trajectory to the human hands for the clip. Finally, we extract six kernel descriptors from the bounding box of each frame in the trajectory: gradient, color, local binary pattern, depth gradient, spin, surface normals, and KPCA/self-similarity, which have been proven to be useful features for \mbox{RGB-D} data~\cite{Wu_2014_RSS}. We concatenate the object features of each frame as the interactive-object-trajectory feature of the clip.

%% file: approach.tex
\section{Learning Model}\label{sec:apc}

%In the model, we take the visual appearance, semantic context and temporal information into account, and simultaneously segment the video and cluster the actions. and infer the forgotten actions.

\begin{figure}[t]
  \begin{center}
  \includegraphics[width=0.8\linewidth]{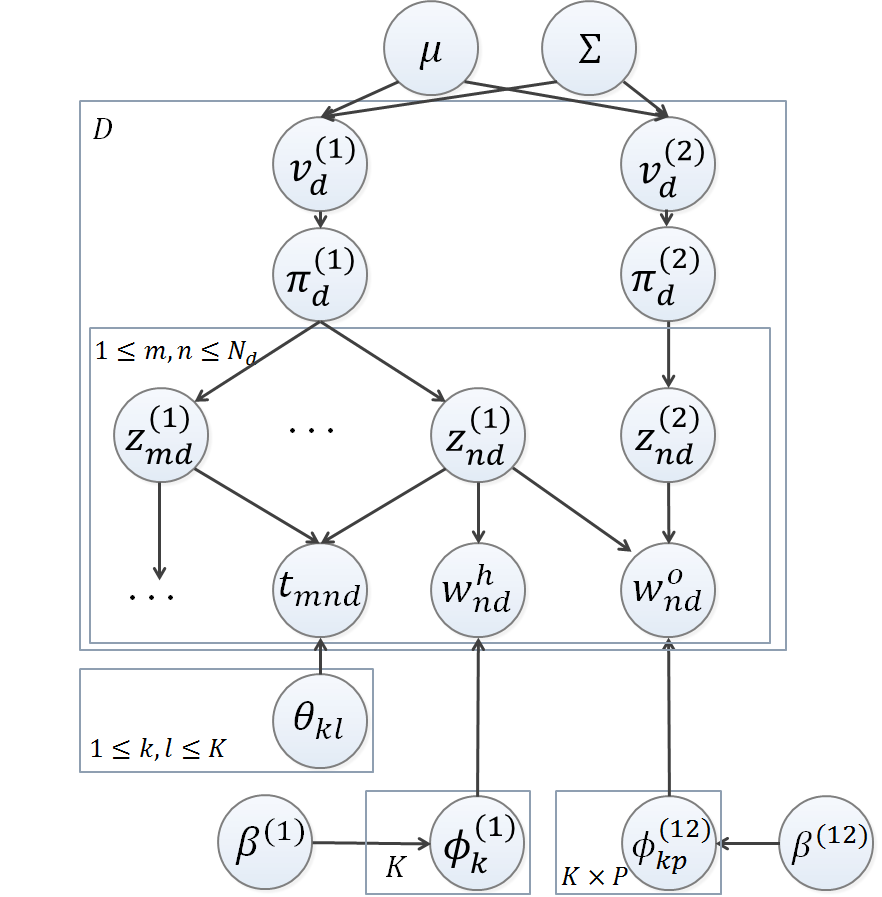}
 \caption{The probabilistic graphic model of our approach.} \label{fig:model}
 \end{center}
\vspace{-0.35in}
\end{figure}

We present a new unsupervised model for our Watch-Bot. The graphic model is illustrated in Fig.~\ref{fig:model} and the notations are in Table~\ref{tb:not}. Our model is able to infer the probability of forgotten actions using the rich relationships between actions and objects.

%In order to capture the rich structure of activities for our robot, we
%present a new generative model based on topic models (see the graphic model in Fig.~\ref{fig:model} and the notations in Table~\ref{tb:not}). The novelty of our model is the ability to infer the probability of forgotten actions by modeling different relationships in a complex activity video.

We learn the model from a training set of $D$ unlabeled videos. Each video as a document $d$ consists of $N_d$ continuous clips $\{c_{nd}\}_{n=1}^{N_d}$, each of which consists of a human-word $w^h_{nd}$ mapped to the human-dictionary and an object-word $w^o_{nd}$ mapped to the object-dictionary. %where $w_{nd}$ is the word index of $n$-th word in $d$-th document.
We assign action-topic to each clip $c_{nd}$ from $K$ latent action-topics, indicating which action-topic they belong to. We assign object-topic to each object-word $w^o_{nd}$ from $P$ latent object-topics, indicating which object-topic is interacted within the clip. The assignments are denoted as $z^{(1)}_{nd}$ and $z^{(2)}_{nd}$. We use superscripts $(1),(2)$ to denote action-topics and object-topics respectively. After assignments, in a video, continuous clips with the same action-topic compose an action segment. All the segments assigned with the same action-topic from the training set compose an action cluster.

As shown in Fig.~\ref{fig:model}, the generative process of our model is as follows. In a document $d$, we choose $z^{(1)}_{dn}\sim Mult(\pi^{(1)}_{:d}), z^{(2)}_{dn}\sim Mult(\pi^{(2)}_{:d})$, where $Mult(\pi)$ is a multinomial distribution with parameter $\pi$. The human-word $w^h_{nd}$ is drawn from an action-topic specific multinomial distribution $\phi^{(1)}_{z^{(1)}_{nd}}$, $w^h_{dn}\sim Mult(\phi^{(1)}_{z^{(1)}_{dn}})$, where $\phi^{(1)}_k\sim Dir(\beta^{(1)})$ is the human-word distribution of action-topic $k$, sampled from a Dirichlet prior with the hyperparameter $\beta^{(1)}$. While the object-word $w^o_{nd}$ is drawn from an action-topic and object-topic specific multinomial distribution $\phi^{(12)}_{z^{(1)}_{nd}z^{(2)}_{nd}}$, $w^o_{dn}\sim Mult(\phi^{(12)}_{z^{(1)}_{nd}z^{(2)}_{nd}})$, where $\phi^{(12)}_{kp}\sim Dir(\beta^{(12)})$ is the object-word distribution of action-topic $k$ and object-topic $p$. Here we consider the same object type like \emph{book} can be variant in appearance in different actions such as a \emph{close book} in \emph{fetch-book} and a \emph{open book} in \emph{reading}. So we consider the object-word distribution for different combinations of the action topic and the object topic.

%{\bf Basic generative process.} In a document $d$, the action-topic assignment $z^{(1)}_{nd}$ is chosen from a multinomial distribution with parameter $\pi^{(1)}_{:d}$, $z^{(1)}_{dn}\sim Mult(\pi^{(1)}_{:d})$, where $\pi^{(1)}_{:d}$ is sampled from a prior.  The same for object-topic assignment $z^{(2)}_{nd}$, $z^{(2)}_{dn}\sim Mult(\pi^{(2)}_{:d})$. The human-word $w^h_{nd}$ is generated by an action-topic specific multinomial distribution $\phi^{(1)}_{z^{(1)}_{nd}}$, $w^h_{dn}\sim Mult(\phi^{(1)}_{z^{(1)}_{dn}})$, where $\phi^{(1)}_k\sim Dir(\beta^{(1)})$ is the human-word distribution of action-topic $k$, sampled from a Dirichlet prior with the hyperparameter $\beta^{(1)}$. While the object-word $w^o_{nd}$ is generated by an action-topic and object-topic specific multinomial distribution $\phi^{(12)}_{z^{(1)}_{nd}z^{(2)}_{nd}}$, $w^o_{dn}\sim Mult(\phi^{(12)}_{z^{(1)}_{nd}z^{(2)}_{nd}})$, where $\phi^{(12)}_{kp}\sim Dir(\beta^{(12)})$ is the object-word distribution of action-topic $k$ and object-topic $p$. Here we consider the same object type like \emph{book} can be variant in appearance in different actions such as a close book in \emph{fetch-book} and a open book in \emph{reading}. So we consider the object-word distribution for different combinations of the action topic and the object topic.

%When we directly use LDA in action videos, we will lose the semantic and temporal relations between action-words, as the topics and topic assignment are all independent.

The co-occurrence such as action \emph{put-down-items} and action \emph{take-items}, object \emph{book} and action \emph{reading}, is useful to recognizing the co-occurring actions/objects and gives a strong evidence for detecting forgotten actions. We model the co-occurrence by drawing their priors from a mixture distribution. In the graphic model, $\pi^{(1)}_{kd}, \pi^{(2)}_{pd}$ decide the probability of action-topic $k$ and object-topic $p$ occurring in a document $d$, where $\sum_{k=1}^K\pi^{(1)}_{kd}=1, \sum_{p=1}^P\pi^{(2)}_{pd}=1$. We construct the probabilities using a stick-breaking process as in~\cite{Wu_2015_CVPR}, where $v^{(1)}_{kd}, v^{(2)}_{pd}$ serve as the priors. Then we draw the packed vector $v_{:d}=[v^{(1)}_{:d}, v^{(2)}_{:d}]$ from a multivariate normal distribution $N(\mu,\Sigma)$%\footnote{In practice, we use a truncated vector $v^{(1)}_{:d}=[v^{(1)}_{1d},\cdots,v^{(1)}_{K-1,d}]$ for (K-1) topics, and set $\pi^{(1)}_{Kd}=1-\sum_{k=1}^{K-1}\pi^{(1)}_{kd}=\prod_{k=1}^{K-1}\Psi(-v^{(1)}_{kd})$ as the probability of the final topic for a valid distribution. The same for $v^{(2)}_{:d}$.}
, which captures the correlations between action-topics and object-topics.

The temporal relations between actions are also useful to discriminating the actions using temporal ordering and inferring the temporal consistent forgotten actions. So we model the relative time of occurring actions as in~\cite{Wu_2015_CVPR}. In detail, let $t_{nd}, t_{md}\in(0,1)$ be the absolute time stamp of $n$-th clip and $m$-th clip, which is normalized by the video length. $t_{mnd}=t_{md}-t_{nd}$ is the relative time of $m$-th clip relative to $n$-th clip. Then $t_{mnd}$ is drawn from a certain distribution, $t_{mnd}\sim \Omega(\theta_{z^{(1)}_{md},z^{(1)}_{nd}})$, where $\theta_{z^{(1)}_{md},z^{(1)}_{nd}}$ are the parameters. $\Omega(\theta_{k,l})$ are $K^{2}$ pairwise action-topic specific relative time distributions defined by a product of a Bernoulli distribution which gives the probability of action $k$ after/before the action $l$, and a normal distribution which estimates how long the action $k$ is after/before the action $l$.

%{\bf Relative time distributions.} Second we model the relative time of occurring actions by taking their time stamps into account. These temporal relations between actions are useful to discriminating the actions using temporal ordering and inferring the temporal consistent forgotten actions.  We consider that the relative time between two clips are drawn from a certain distribution according to their action-topic assignments.  In detail, let $t_{nd}, t_{md}\in(0,1)$ be the absolute time stamp of $n$-th clip and $m$-th clip, which is normalized by the video length. $t_{mnd}=t_{md}-t_{nd}$ is the relative time of $m$-th clip relative to $n$-th clip. Then $t_{mnd}$ is drawn from a certain distribution, $t_{mnd}\sim \Omega(\theta_{z^{(1)}_{md},z^{(1)}_{nd}})$, where $\theta_{z^{(1)}_{md},z^{(1)}_{nd}}$ are the parameters. $\Omega(\theta_{k,l})$ are $K^{2}$ pairwise action-topic specific relative time distributions defined as follows:

\if 0
\begin{equation}\label{eqn:tpdf}
\begin{split}
\Omega(t|\theta_{k,l})&=
\begin{cases}
b_{k,l}\cdot N(t|\theta^+_{k,l}) \ \ \ \ \ \ \ \ \textrm{if}\ \ t\geq 0,\\		
1-b_{k,l}\cdot N(t|\theta^-_{k,l}) \ \ \ \textrm{if}\ \ t<0,\\
\end{cases}
\end{split}
\end{equation}
\fi

 \begin{table}[t]
\footnotesize
\setlength{\tabcolsep}{2pt}
%\begin{center}
\caption{Notations in our model.}\label{tb:not}
\vspace{-0.1in}
\begin{tabular}{ll}
\hline
Symbols&Meaning\\
\hline
$D$ &number of videos in the training database;\\
$K$ &number of action-topics;\\
$P$ &number of object-topics;\\
$N_d$ &number of human-words/object-words in a video;\\
$c_{nd}$ &$n$-th clip in $d$-th video;\\
$w^h_{nd}$ &$n$-th human-word in $d$-th video;\\
$w^o_{nd}$ &$n$-th object-word in $d$-th video;\\
$z^{(1)}_{nd}$ &action-topic assignment of $c_{nd}$;\\
$z^{(2)}_{nd}$ &object-topic assignment of $w^o_{nd}$;\\
$t_{nd}$ &normalized timestamp of of $c_{nd}$;\\
$t_{mnd}$&$=t_{md}-t_{nd}$ the relative time between $c_{md}$ and $c_{nd}$;\\
$\pi^{(1)}_{:d}, \pi^{(2)}_{:d}$ &the probabilities of action/object-topics in $d$-th document;\\
$v^{(1)}_{:d}, v^{(2)}_{:d}$ &the priors of $\pi^{(1)}_{:d}, \pi^{(2)}_{:d}$ in $d$-th document;\\
$\phi^{(1)}_k$ &multinomial human-word distribution from action-topic $k$;\\
$\phi^{(12)}_{kp}$ &multinomial object-word distribution from \\
&action-topic $k$ and object-topic $p$;\\
$\mu,\Sigma$ &multivariate normal distribution of $v_{:d}=[v^{(1)}_{:d}, v^{(2)}_{:d}]$;\\
$\theta_{kl}$ &relative time distribution of $t_{mnd}$, between action-topic $k,l$;\\
\hline
\end{tabular}
\vspace{-0.2in}
\end{table}

%% file: learning.tex
\subsection{Learning and Inference}\label{sec:learn}
We use Gibbs sampling~\cite{Blei_2009_TM,Kim_2011_NIPS} to learn the parameters and the infer the hidden variables from the posterior distribution of our model. The word $w^h_{nd}, w^o_{hd}$ and the relative time $t_{mnd}$ are observed in each video. We can integrate out $\Phi^{(1)}_k,\Phi^{(12)}_k$ since $Dir(\beta^{(1)}), Dir(\beta^{(12)})$ are conjugate priors for the multinomial distributions $\Phi^{(1)}_k, \Phi^{(12)}_k$. We also estimate the standard distributions including the mutivariate normal distribution $N(\mathbf{\mu},\Sigma)$ and the time distribution $\Omega(\theta_{kl})$ using the method of moments, once per iteration of Gibbs sampling. The topic priors $v^{(1)}_{:d}, v^{(2)}_{:d}$ can be sampled by a Metropolis-Hastings independence sampler~\cite{Gelman_2013_bayesian} as in~\cite{Wu_2015_CVPR}. Following the convention, we use the fixed symmetric Dirichlet distributions by setting $\beta^{(1)}, \beta^{(12)}$ as $0.01$.

%Gibbs sampling is commonly used as a means of statistical inference to approximate the distributions of variables when direct sampling is difficult~\cite{Blei_2009_TM,Kim_2011_NIPS}. Given a video, the word $w^h_{nd}, w^o_{hd}$ and the relative time $t_{mnd}$ are observed. Given the training videos, we use Gibbs sampling to approximately sample other hidden variables from the posterior distribution of our model. Since we adopt conjugate prior $Dir(\beta^{(1)}), Dir(\beta^{(12)})$ for the multinomial distributions $\Phi^{(1)}_k, \Phi^{(12)}_k$, we can integrate out $\Phi^{(1)}_k,\Phi^{(12)}_k$ and need not to sample them. For simplicity and efficiency, we estimate the standard distributions including the mutivariate normal distribution $N(\mathbf{\mu},\Sigma)$ and the time distribution $\Omega(\theta_{kl})$ by the method of moments, once per iteration of Gibbs sampling. Similar to many applications using the topic model, we use fixed symmetric Dirichlet distributions by setting $\beta^{(1)}, \beta^{(12)}$ as $0.01$.

Then we introduce how we sample the topic assignment $z^{(1)}_{nd}, z^{(2)}_{nd}$. We do a collapsed sampling as in Latent Dirichlet Allocation (LDA)~\cite{Blei_2003_LDA} by calculating the posterior distribution of $z^{(1)}_{nd}, z^{(2)}_{nd}$:
\vspace{-0.02in}
\begin{align}\label{eqn:posz}
&p(z^{(1)}_{nd}=k|\pi^{(1)}_{:d},z^{(1)}_{-nd},z^{(2)}_{nd},t_{nd})\notag\\
&\propto \pi^{(1)}_{kd}\omega(k,w^h_{nd})\omega(k,z^{(2)}_{nd},w^o_{nd})p(t_{nd}|z^{(1)}_{:d},\theta),\notag\\
&p(z^{(2)}_{nd}=p|\pi^{(2)}_{:d},z^{(2)}_{-nd},z^{(1)}_{nd})\propto
\pi^{(2)}_{pd}\omega(z^{(1)}_{nd},p,w^o_{nd}),\notag\\
&\omega(k,w^h_{nd})=\frac{N_{kw^h}^{-nd}+\beta^{(1)}}{N_{k}^{-nd}+N_w\beta^{(1)}},\notag\\
&\omega(k,p,w^o_{nd})=\frac{N_{kpw^o}^{-nd}+\beta^{(12)}}{N_{kp}^{-nd}+N_o\beta^{(12)}},\notag\\
&p(t_{nd}|z^{(1)}_{:d},\theta)=\prod_m^{N_d} \Omega(t_{mnd}|\theta_{z^{(1)}_{md},k})\Omega(t_{nmd}|\theta_{k,z^{(1)}_{md}}),
\vspace{-0.05in}
\end{align}
where $N_w,N_o$ is the number of unique word types in dictionary, $N_{kw^h}^{-nd}/N_{kpw^o}^{-nd}$ denotes the number of instances of word $w^h_{nd}/w^o_{nd}$ assigned with action-topic $k$/action-topic $k$ and object-topic $p$, excluding $n$-th word in $d$-th document, and $N_{k}^{-nd}/N_{kp}^{-nd}$ denotes the number of total words assigned with action-topic $k$/action-topic $k$ and object-topic $p$. $z^{(1)}_{-nd}/z^{(2)}_{-nd}$ denotes the topic assignments for all words except $z^{(1)}_{nd}/z^{(2)}_{nd}$.

In Eq.~(\ref{eqn:posz}), note that the topic assignments are decided by which actions/objects are more likely to co-occur in the video (the occurance probabilities $\pi^{(1)}_{kd}/\pi^{(2)}_{kd}$), the visual appearance of the word (the word distributions $\omega(k,w^h_{nd}), \omega(k,p,w^o_{nd})$) and the temporal relations (the relative time distributions $p(t_{nd}|z^{(1)}_{:d},\theta)$). The time complexity of the sampling per iteration is $O(N_dD(\max(N_dK,P))$. 

%$\pi^{(1)}_{kd}/\pi^{(2)}_{kd}$ is the topic prior generated by a joint distribution giving which actions/objects are more likely to co-occur in the video. $\omega(k,w^h_{nd})$ is the word distribution for action-topic $k$ giving which action-topic the clip is more likely from. $\omega(k,p,w^o_{nd})$ is the word distribution for action-topic $k$ and object-topic $p$ giving which action-topic and object-topic the object word is more likely from. $p(t_{nd}|z^{(1)}_{:d},\theta)$ is the time distribution giving which action-topic assignment is more causally consistent to other action-topic assignments.

For inference of a test video, we sample the unknown topic assignments $z^{(1)}_{nd}, z^{(2)}_{nd}$ and the topic priors $v^{(1)}_{:d},v^{(2)}_{:d}$ using the learned parameters in the training stage.

%\section{Action Segmentation and Recognition}
 %\vspace{-0.05in}
%After we learn the topic-assignment of each action-word, we can easily get the action segments by merging the continuous clips with the same assigned topic. Also the assigned topic of the segment indicate which action it is and these segments with the same assigned topic consist an action-topic segment cluster.

%% file: patching.tex
% !TEX root = patch.tex

{\setlength{\textfloatsep}{2pt}
\begin{algorithm}[t]
\small
\caption{Forgotten Action and Object Detection.}
\begin{algorithmic}
\STATE \textbf{Input:} RGB-D video $q$ with tracked human skeletons.
\STATE \textbf{Output:} Claim no action forgotten, or output an action segment with the forgotten action and a bounding box of the related object in the current scene.
\STATE 1. Assign the action-topics to clips and the object-topics to object-words in $q$ as introduced in Section~\ref{sec:learn}.
\STATE 2. Get the action segments by merging the continuous clips with the same assigned action-topic.
\STATE 3. If the assigned action-topics $K_e$ in $q$ contains all modeled action-topics $[1:K]$, claim no action forgotten and return;
\STATE 4. For each action segmentation point $t_s$, each not assigned action-topic $k_m\in{[1:K]}-K_e$, and each object-topic $p_m\in{[1:P]}$:
\STATE \ \ \ Compute the probability defined in Eq.~\ref{eqn:patch};
\STATE 5. Select the top tree possible tuples $(k_m,p_m,t_s)$, and get the forgotten action segment candidate set $Q$ which contains segments with topics $(k_m,p_m)$;
\STATE 6. Select the top forgotten action segment $p$ from $Q$ with the maximum $forget\_score(p)$;
\STATE 7. If $forget\_score(p)$ is smaller than a threshold, claim no action forgotten and return;
\STATE 8. Segment the current frame to super-pixels using edge detection~\cite{Dollar_2013_ICCV} as in Section~\ref{sec:ov};
\STATE 9. Select the nearest super-pixels to both extracted object bounding box in $q$ and $p$.
\STATE 10. Merge the adjacent super-pixels and bound the largest one with a rectangle as the output bounding box.
\STATE 11. Return the top forgotten action segment and the object bounding box.
\end{algorithmic}  \label{alg:adr}
\end{algorithm}

 \section{Forgotten Action Detection and Reminding}\label{sec:ap}

In this section, we describe how we apply our model in our robot to detecting the forgotten actions and reminding people. It is more challenging than conventional action recognition, since what to infer is not shown in the query video. Therefore, unlike the existing models on action relations learning, our model learns rich relations rather than the only local temporal transitions. As a result, those actions occurred with a relatively large time interval, occurred after the forgotten actions, as well as the interactive objects can also be used to detect forgotten actions, \eg, a \emph{put-back-book} might be forgotten as previously seen a \emph{fetch-book} action before a long \emph{reading}, and seen a \emph{book} and a \emph{leaving} action indicates he really forgot it.

Our goal is to detect the forgotten action and then point out the related object in the forgotten action using our learned model (see Alg.~\ref{alg:adr}). We first use our model to segment the query video into action segments (step 1,2 in Alg.~\ref{alg:adr}), and then infer the most possible forgotten action-topic and the related object-topic (step 4 in Alg.~\ref{alg:adr}). Next we retrieve a top forgotten action segment from the training database, containing the inferred forgotten action-topic and the object-topic (step 5,6 in Alg.~\ref{alg:adr}). Using the extracted object in the retrieved segment, we detect the bounding box of the related forgotten object in the Kinect's view of the query video (step 8,9,10 in Alg.~\ref{alg:adr}). After that, we map the bounding box of the object from the Kinect's view to the camera's view. Finally, we pan/tilt camera until its laser pointer points out the related object in the current scene.

%After we learn the action-topic assignment of each clip, we can easily get the action segments by merging the continuous clips with the same assigned action-topic. Also the assigned action-topic of the segment indicate which action it is and these segments with the same assigned action-topic consist an action-topic segment cluster, that can be used for inferring which action-topic is forgotten. Moreover, we can get which object-topics are in each action segment according to the object-topic assignments.

\begin{figure}[t]
  \begin{center}
  \includegraphics[width=0.9\linewidth]{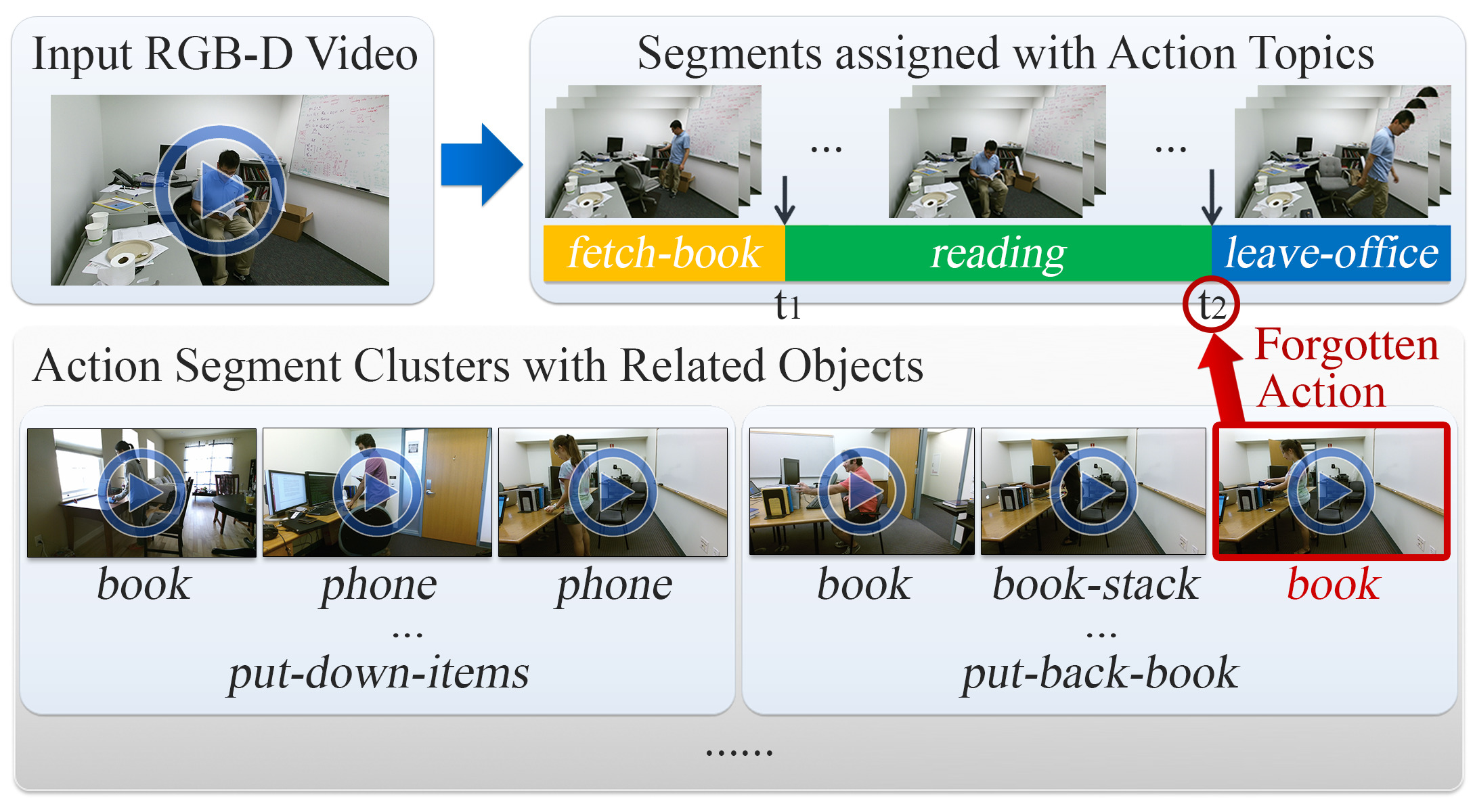}
  \caption{{\bf Illustration of forgotten action and object detection using our model.} Given a query video, we infer the forgotten action-topic and object-topic in each segmentation point ($t_1,t_2$). Then we select the top segment from the inferred action-topic's segment cluster with the inferred object-topic with the maximum $forget\_score$.}
   \label{fig:vp}
 \end{center}
\vspace{-0.1in}
\end{figure}

{\bf Forgotten Action and Object Inference.}  We first introduce how we infer the forgotten action-topic and object-topic using the dependencies in our learned model. After assigning the action-topics and object-topics to the query video $q$, we consider adding one additional clip $\hat{c}$ consisting of $\hat{w^h}, \hat{w^o}$ into $q$ in every action segmentation point $t_s$ (see Fig~\ref{fig:vp}). Then the probabilities of the missing action-topics $k_m$ with object-topics $p_m$ in each segmentation point $t_s$ can be computed following the posterior distribution in Eq.~(\ref{eqn:posz}):

%Our model infers the forgotten action using the probability inference based on the dependencies between actions and objects. After assigning the action-topics and object-topics to a query video $q$, we consider adding one additional clip $\hat{c}$ consisting of $\hat{w^h}, \hat{w^o}$ into $q$ in each action segmentation point $t_s$ (see Fig~\ref{fig:vp}). Then the probabilities of the missing action-topics $k_m$ with object-topics $p_m$ in each segmentation point $t_s$ can be compared following the posterior distribution in Eq.~(\ref{eqn:posz}):
\vspace{-0.1in}
\begin{align}\label{eqn:patch}
&p(z^{(1)}_{\hat{c}}=k_m, z^{(2)}_{\hat{c}}=p_m, t_{\hat{c}}=t_s|other)\notag\\
&\propto \pi^{(1)}_{k_md}\pi^{(2)}_{p_md}p(t_{s}|z^{(1)}_{:d},\theta)\sum_{w^h,w^o}\omega(k_m,w^{h})\omega(k_m,p_m,w^{o}),\notag\\
&\st \ \ \ t_s\in{T_s},\ k_m\in{[1:K]}-K_e,
\end{align}
where $T_s$ is the set of segmentation points (such as $t_1,t_2$ in Fig.~\ref{fig:vp}) and $K_e$ is the set of existing action-topics in the video (\emph{fetch-book}, \etc in Fig.~\ref{fig:vp}). Thus ${[1:K]}-K_e$ are the missing topics in the video (\emph{put-down-items}, \etc in Fig.~\ref{fig:vp}). $p(t_{s}|z^{(1)}_{:d},\theta),\omega(k_m,w^{h}),\omega(k_m,p_m,w^{o})$ can be computed as in Eq.~(\ref{eqn:posz}). Here we marginized $\hat{w^h},\hat{w^o}$ to avoid the effect of a specific human-word or object-word. 

Note that, in Eq.~(\ref{eqn:patch}), the closer topics would have higher probabilities $\pi^{(1)}_{kd}, \pi^{(2)}_{pd}$ to co-occur in this query video as they are drawn from the learned joint distribution. The action-topics which are more consistent with the learned temporal relations would have higher probability $p(t_{s}|z^{(1)}_{:d},\theta)$. The marginalized word-topic distribution $\sum_{w^h,w^o}\omega(k_m,w^{h})\omega(k_m,p_m,w^{o})$ give the likelihood of the topic learned from training data.

 % $\pi^{(1)}_{kd}, \pi^{(2)}_{pd}$ gives the probability of a missing action-topic with an object-topic in the video decided by the correlation we learned in the joint distribution prior, \ie, the close topics have higher probabilities to occur in this query video. And $p(t_{s}|z^{(1)}_{:d},\theta)$ measures the temporal consistency of adding a new action-topic. And the marginalized word-topic distribution $\sum_{w^h,w^o}\omega(k_m,w^{h})\omega(k_m,p_m,w^{o})$ give the likelihood of the topic learned from training data.

{\bf Forgotten Action and Object Detection.} We then introduce how we retrieve a top action segment from the training database. We first select the top three tuples $(k_m,p_m,t_s)$ using the above probability. These action segments consist a forgotten action candidate segment set $Q$. We then retrieve the segment from $Q$ with the maximum $forget\_score(p) = ave(\D(f_{pm},f_{qf}),\D(f_{pm},f_{qt}))-max(\D(f_{pf},f_{qt}),\D(f_{pt},f_{qf}))$, where $\D(,)$ is the average pairwise distances between frames, $ave(,),max(,)$ are the average and max value. The front and the tail of the forgotten action segment $f_{pf}, f_{pt}$ need to be similar to the tail of the adjacent segment in $q$ before $t_s$ and the front of the adjacent segment in $q$ after $t_s$: $f_{qt}, f_{qf}$. The middle of the forgotten action segment $f_{pm}$ need to be different to $f_{qt},f_{qf}$, as it is a different action forgotten in the video\footnote{Here the middle, front, tail frames are $20\%$-length of segment centering on the middle frame, starting from the first frame, and ending at the last frame in the segment respectively.}. If the maximum score is below a threshold or there is no missing topics $(\ie, K_e=[1:K])$ in the query video, we claim there is no forgotten actions.

%Then we select the top three tuples $(k_m,p_m,t_s)$ using the above probability. The action segments of action-topic $k_m$ containing object-topic $p_m$ in the training set consist a patched action candidate segment set $Q$. We then select the patched action segment from $Q$ with the maximum $patch\_score$ defined in Eq.~\ref{eqn:rank}. In detail, we consider that the front and the tail of the patched action segment $f_{pf}, f_{pt}$ should be similar to the tail of the adjacent segment in $q$ before $t_s$ and the front of the adjacent segment in $q$ after $t_s$: $f_{qt}, f_{qf}$. At the same time, the middle of the patched action segment $f_{pm}$ should be different to $f_{qt},f_{qf}$, as it is a different action forgotten in the video.\footnote{Here the middle, front, tail frames are $20\%$-length of segment centering on the middle frame, starting from the first frame, and ending at the last frame in the segment respectively.}
\if 0
\begin{equation}\label{eqn:rank}
\begin{split}
forget\_score(p) = ave(\D(f_{pm},f_{qf}),\D(f_{pm},f_{qt}))\\
-max(\D(f_{pf},f_{qt}),\D(f_{pt},f_{qf})),
\end{split}
\end{equation}
\fi
%where $\D(,)$ is the average pairwise distances between frames, $ave(,),max(,)$ are the average and max value. If the maximum score is below a threshold or there is no missing topics $(\ie, K_e=[1:K])$ in the query video, we claim there is no forgotten actions. 

Then we detect the bounding box of the related forgotten object in the current scene. We segment the current frame into super-pixels as in Section~\ref{sec:ov}, then search the nearest super-pixels using the extracted object in the top retrieved action, finally merge the adjacent super-pixels and bound the largest one with a bounding box.

{\bf Real Object Pointing.} We describe how we pan/tilt the camera to point out the real object. We first compute the transformation homography matrix between the frame of the Kinect and the frame of the pan/tilt camera using keypoints matching and RANSAC, which can be done very fast within $0.1$ second. Then we can transform the detected bounding box from the Kinect's view to the pan/tilt camera's view. Since we fix the position of the laser spot in the pan/tilt camera view, next we only need to pan/tilt the camera till the laser spot lies within the bounding box of the target object. To avoid the coordinating error caused by distortion and inconsistency of the camera movement, we use an iterative search plus small step movement instead of one step movement to localize the object (illustrated in Fig.~\ref{fig:system}). In each iteration, the camera pan/tilt a small step towards to the target object according to the relative position between the laser spot and the bounding box. Then the homography matrix is recomputed in the new camera view, so that the bounding box is mapped in the new view. Until the laser spot is close enough to the center of the bounding box, the camera stops moving.

%% file: experiments.tex
% !TEX root = patch.tex

\section{Experiments}\label{sec:exp}

\subsection{Dataset}
\begin{figure}[t]
\begin{center}

  \subfigure[turn-off-monitor]{
  \begin{minipage}{0.46\linewidth}
  \includegraphics[width=1\linewidth]{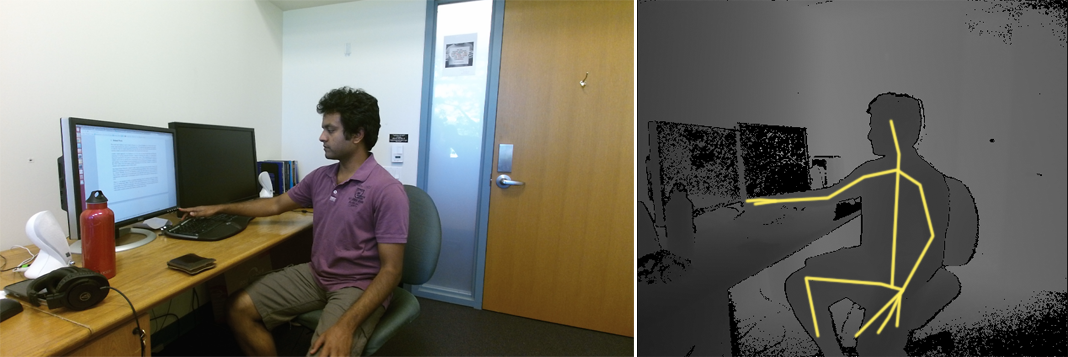}\vspace{0.01in}
  \end{minipage}
  }\hspace{-0.05in}
  \vspace{-0.05in}
  \subfigure[take-item]{
  \begin{minipage}{0.46\linewidth}
  \includegraphics[width=1\linewidth]{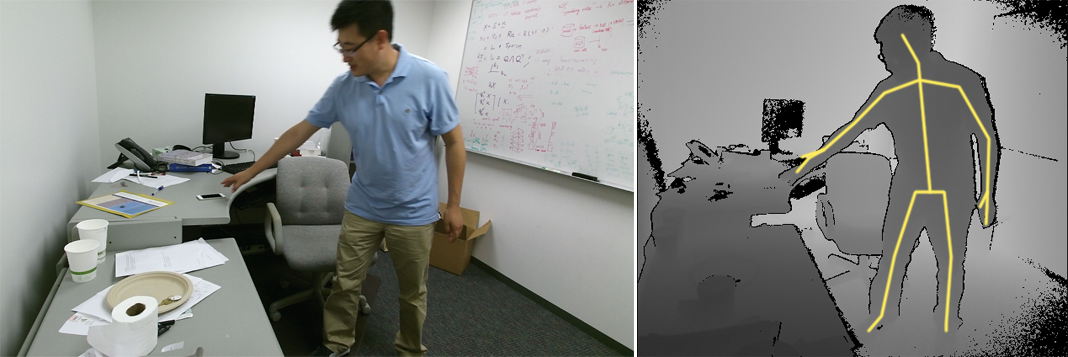}\vspace{0.01in}
  \end{minipage}
  }\hspace{-0.05in}
	
\subfigure[fetch-from-fridge]{
  \begin{minipage}{0.46\linewidth}
  \includegraphics[width=1\linewidth]{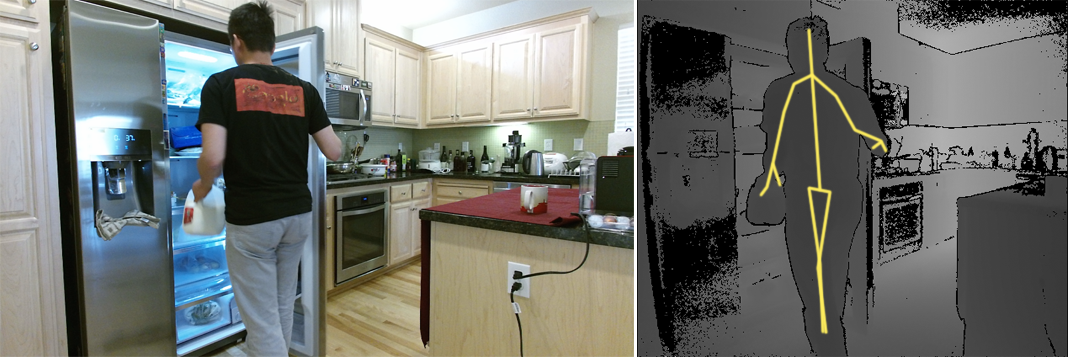}\vspace{0.01in}
  \end{minipage}
  }\hspace{-0.05in}
\subfigure[fill-kettle]{
  \begin{minipage}{0.46\linewidth}
  \includegraphics[width=1\linewidth]{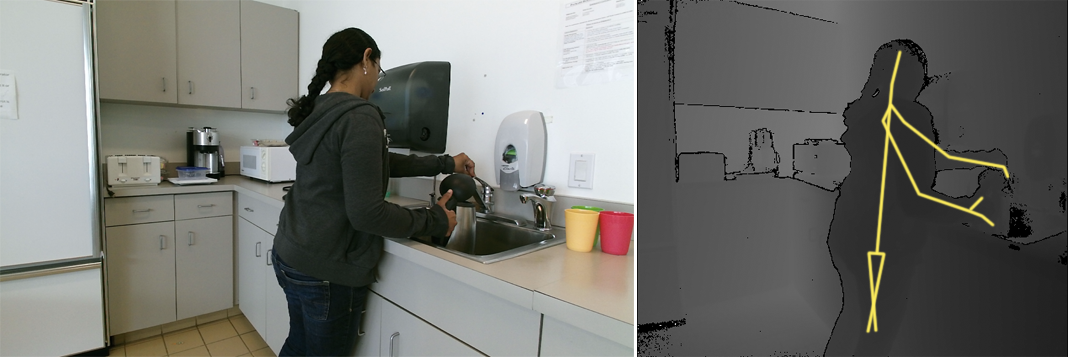}\vspace{0.01in}
  \end{minipage}
  }\hspace{-0.05in}
  \vspace{-0.1in}
  	 \caption{Action examples in the dataset. The left is RGB frame and the right is depth frame with human skeleton (yellow).}
   \label{fig:actions}
 \end{center}
\vspace{-0.2in}
\end{figure}

\vspace{-0.05in}
We evaluate our Watch-Bot in a challenging human activity \mbox{RGB-D} dataset~\cite{Wu_2015_CVPR} consisting of $458$ videos of about $230$ minutes in total recorded by the Kinect v2 sensor. Each video in the dataset contains $2$-$7$ actions interacted with different objects (see examples in Fig.~\ref{fig:actions}). We asked $7$ subjects to perform human daily activities in $8$ offices and $5$ kitchens with complex backgrounds and recorded the activities in different views. It is composed of fully annotated $21$ types of actions ($10$ in the office, $11$ in the kitchen) interacted with $23$ types of objects. The participants finish tasks with different combinations of actions and ordering. Some actions occur together often such as \emph{fill-kettle} and \emph{boil-water}, while some are not always together. Some actions are in a fix order such as \emph{turn-on-monitor} and \emph{turn-off-monitor} while some occur in random order. Also, in the dataset, people forgot actions in $222$ videos. There are $3$ types of forgotten actions in 'office' and $5$ types in 'kitchen'. %Moreover, to evaluate the action patching performance,  %Besides forgetting those high-level actions we labeled, we also provide some cases with sub-action forgotten such as \emph{leave-kitchen(take cup)} and \emph{fill-kettle(turn-off-tap)}, where forgotten sub-action is in the brackets.

\vspace{-0.05in}
\subsection{Baselines}
\vspace{-0.05in}
We compare four unsupervised approaches. They are Hidden Markov Model (HMM)~\cite{Baum_1966_HMM}, LDA topic model~\cite{Blei_2003_LDA}, our previous work Causal Topic Model(CaTM)~\cite{Wu_2015_CVPR} and our Watch-Bot Topic Model (WBTM). We use the same human skeleton and \mbox{RGB-D} features introduced in Section~\ref{sec:ov}. In LDA, actions and objects are modeled independently as the priors of action/object assignments are sampled from a fix Dirichlet prior and there is no relative time between actions modeled. For HMM, similarly we set action states which generates both human and object trajectory features of each clip, and object states which generates object trajectory features. Since there is no object modeled in CaTM, we only evaluate its activity related performance.

%We also evaluate HMM and our model CaTM using the popular features for action recognition, dense trajectories feature (DTF)~\cite{Wang_2011_CVPR}, extracted only in RGB videos\footnote{We train a codebook with the size of $2000$ and encode the extracted DTF features in each clip as the bag of features using the codebook.}, named as HMM-DTF and CaTM-DTF.

In the experiments, we set the number of action-topics/object-topicss and states for HMM equal to or more than ground-truth action/object classes. For LDA, CaTM and our WBTM, the clip length is set to $20$ frames, densely sampled by step one and the size of human/object dictionary is set to $500$. The forgotten action candidate set for different approaches consists of the segments with the inferred missing topics by transition probabilities for HMM, the topic priors for LDA. After inference, we use the same forgotten action and object detection method as introduced in Section~\ref{sec:ap}.

\subsection{Evaluation Metrics}
We test in two environments `office' and `kitchen'. In each environment, the dataset is split into a train set with most full videos (office: $87$, kitchen $119$) and a few forgotten videos (office: $10$, kitchen $10$), and a test set with a few full videos (office: $10$, kitchen $20$) and most forgotten videos (office: $89$, kitchen $113$). We train the models in the train set and evaluate the following metrics in the test set.

%In each environment, we split the data into a train set with most full videos (office: $90$, kitchen $110$) and a few forgotten videos (office: $10$, kitchen $20$), and a test set with a few full videos (office: $10$, kitchen $10$) and most forgotten videos (office: $90$, kitchen $120$).
%We evaluate the unsupervised action segmentation and recognition by comparing obtained topics of each segment by models with the ground truth segment action labels. In the experiments, we set the topic numbers larger than (or equal to) the ground truth action numbers, since topics could be mapped to sub-actions within a semantic action.

{\bf Action Segmentation and Cluster Assignment.} As in evaluation for unsupervised clustering, we map the action cluster in the train set to the ground-truth action labels by counting the mapped frames between action-topics and ground-truth action classes as in~\cite{Wu_2015_CVPR} . Then we can use the mapped action class label for evaluation.

We measure the performance in two ways. Per frame: we compute \emph{frame-wise accuracy (Frame-Acc)}, the ratio of correctly labeled frames. Segmentation: we consider a true positive if the union/intersection of the detected and the ground-truth segments is greater than $40\%$ as in~\cite{Pirsiavash_2014_CVPR}. We compute \emph{segmentation accuracy (Seg-Acc)}, the ratio of the ground-truth segments that are correctly detected and \emph{segmentation average precision (Seg-AP)} by sorting all action segments using the average probability of their words'  topic assignments. All above three metrics are computed by taking the average of each action class.

{\bf Forgotten Action and Object Detection.} We measure the \emph{forgotten action detection accuracy (FA-Acc)} by the portion of correct detected forgotten action or correctly claiming no forgotten actions. We consider the output forgotten action segments by the compared approaches containing over $50\%$ ground-truth forgotten actions as correct. We measure the \emph{forgotten object detection accuracy (FO-Acc)} by the typical object detection metric,  that considers a true positive if the overlap rate (union/intersection) between the detected and the ground-truth object bounding box is greater than $40\%$.

\begin{table}[t]
\footnotesize
\setlength{\tabcolsep}{6pt}
\begin{center}
\caption{Action segmentation and cluster assignment results, and forgotten action/object detection results.}\label{tb:re}
\begin{tabular*}{\linewidth}{@{\extracolsep{\fill}}c|c c c c c}
\hline
`office'(\%)&Seg-Acc&Seg-AP&Frame-Acc&FA-Acc&FO-Acc\\
%(\%)&Offline&Online&Offline&Online&Offline&Online\\
\hline\hline
HMM&19.4&23.1&27.3&32.2&20.4\\
LDA&12.2&19.6&18.4&15.7&10.5\\
CaTM&32.9&34.6&38.5&41.5&-\\
WBTM&\textbf{35.2}&\textbf{36.0}&\textbf{41.2}&\textbf{46.2}&\textbf{36.4}\\
\hline
\hline
`kitchen'(\%)&Seg-Acc&Seg-AP&Frame-Acc&FA-Acc&FO-Acc\\
%(\%)&Offline&Online&Offline&Online&Offline&Online\\
\hline\hline
HMM&17.2&18.8&20.3&12.4&5.3\\
LDA&6.7&17.1&14.4&10.8&5.3\\
CaTM&29.0&25.5&34.0&20.5&-\\
WBTM&\textbf{30.7}&\textbf{28.5}&\textbf{36.9}&\textbf{24.4}&\textbf{20.6}\\
\hline
\end{tabular*}
\end{center}
\vspace{-0.1in}
\end{table}

%Then we use three metrics to measure the performance. First is the average class frame-wise accuracy, which is computed for each class $c$ as follows:
%\begin{equation}
%AC(c)=\frac{\sum_i \delta(map(k_i),c)}{\sum_i\delta(c_i,c)}.\notag
%\end{equation}

%To further evaluate the smoothness of the action segmentation. We further evaluate the segmentation accuracy and average precision. We consider a true positive if the overlap (union/intersection) between the detected segmentation and the ground-truth action segment is more than a default threshold $40\%$~\cite{Pirsiavash_2014_CVPR}. Then segmentation accuracy and average precision are computed as follows:

%\begin{equation}
%S-AC(c)=\frac{N_{gc}}{N_c},\ SP(c)=\frac{N_{tc}}{N_{pc}},\notag
%\end{equation}
%where $N_{gc}$ is the number of ground-truth segments of class $c$, which have a corresponding true positive detected segmentation, and $N_c$ is the number of ground-truth segments of class $c$. $N_{tc}$ is the number of true positive detected segments of class $c$ and $N_{pc}$ is the number of the detected segments assigned with $map(k)=c$.

\vspace{-0.05in}
\subsection{Results}
\vspace{-0.05in}

\begin{figure}[t]
  \begin{center}\hspace{-0.3in}
  \includegraphics[width=0.54\linewidth]{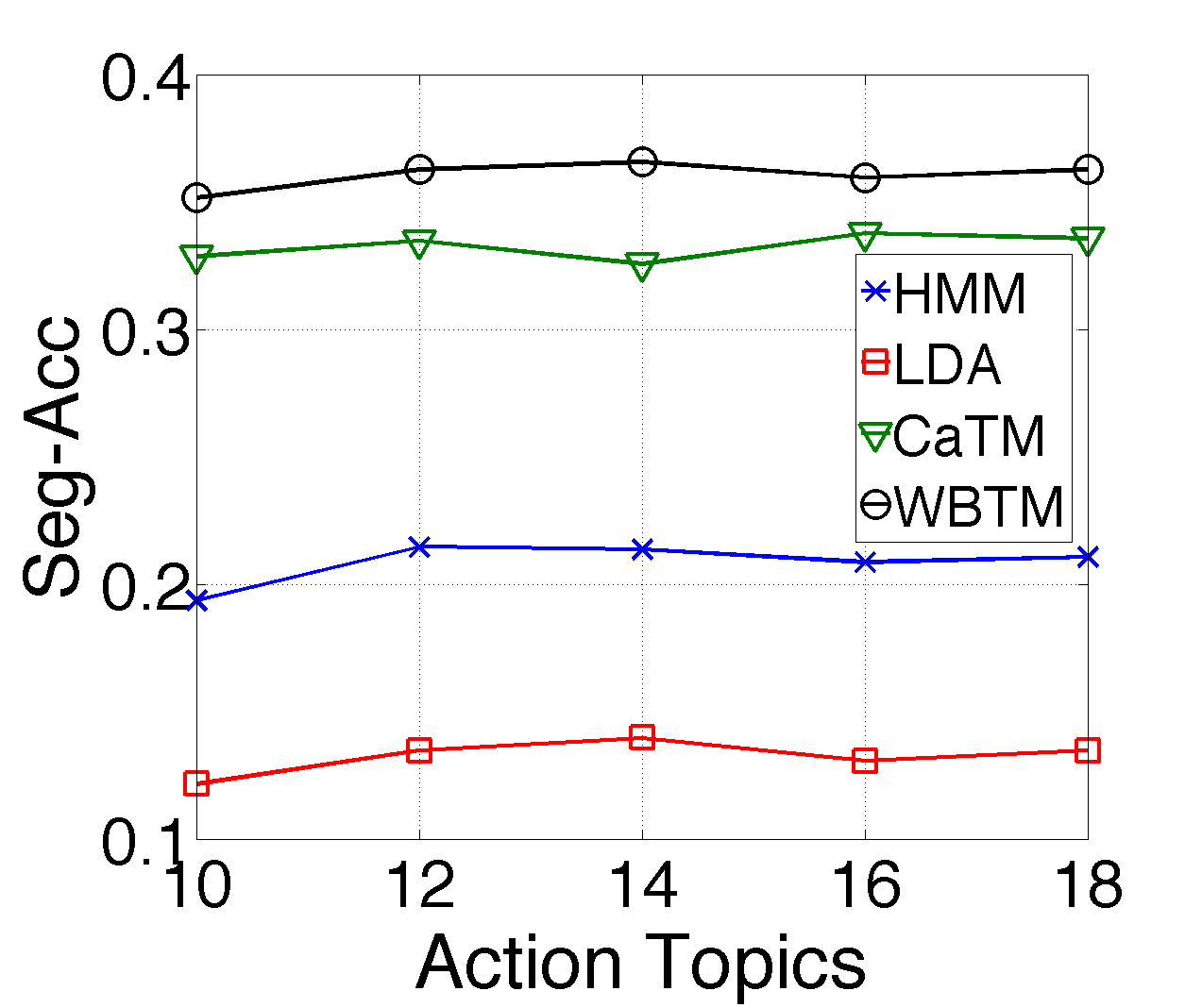}\hspace{-0.15in}
  \includegraphics[width=0.54\linewidth]{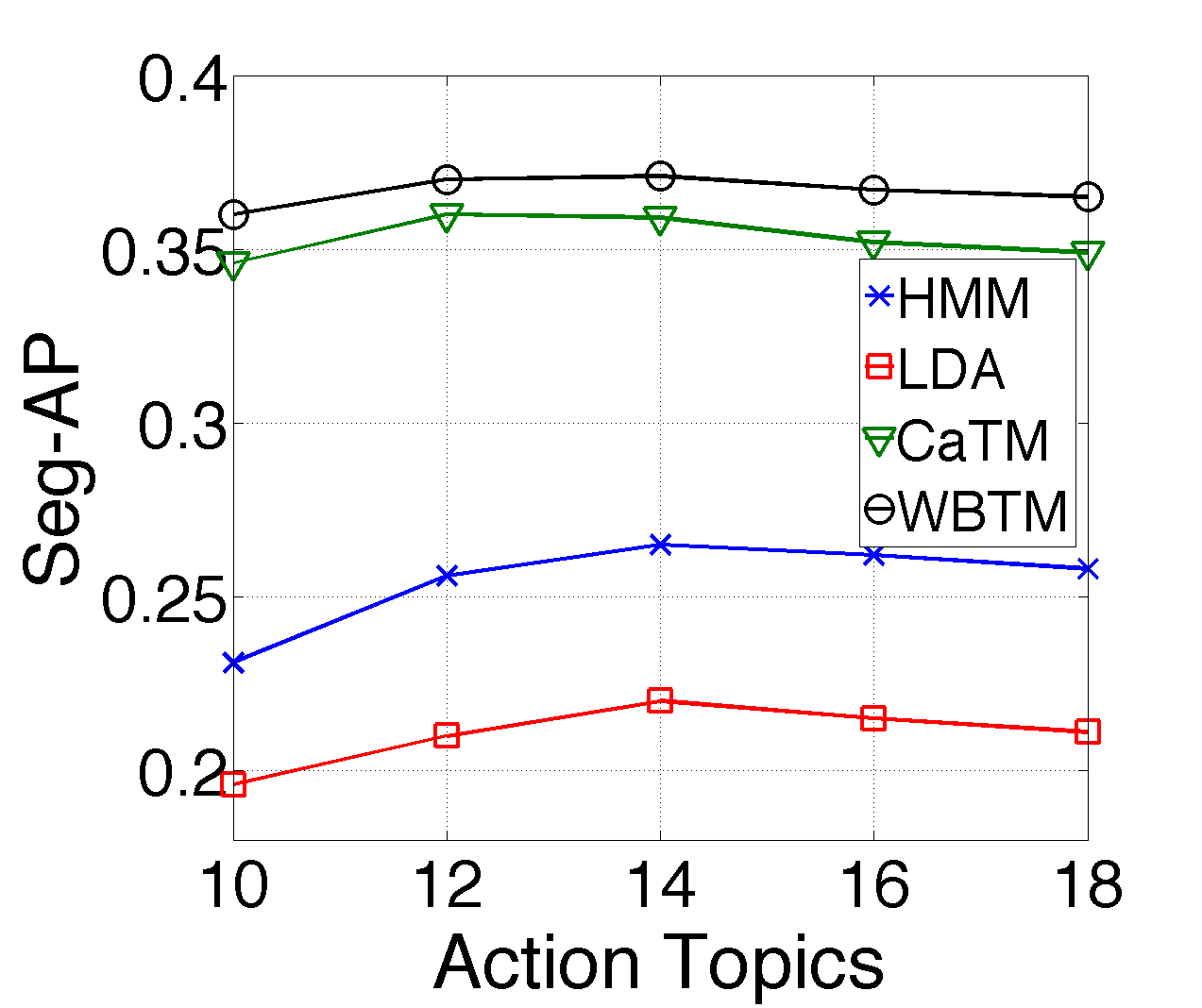}\hspace{-0.3in}
  \hspace{-0.2in}
  \vspace{-0.1in}
 \caption{Action segmentation Acc/AP varied with the number of action-topics in `office' dataset.} \label{fig:act}
 \end{center}
  \vspace{-0.25in}
\end{figure}

\begin{figure}[t]
  \begin{center}\hspace{-0.3in}
  \includegraphics[width=0.54\linewidth]{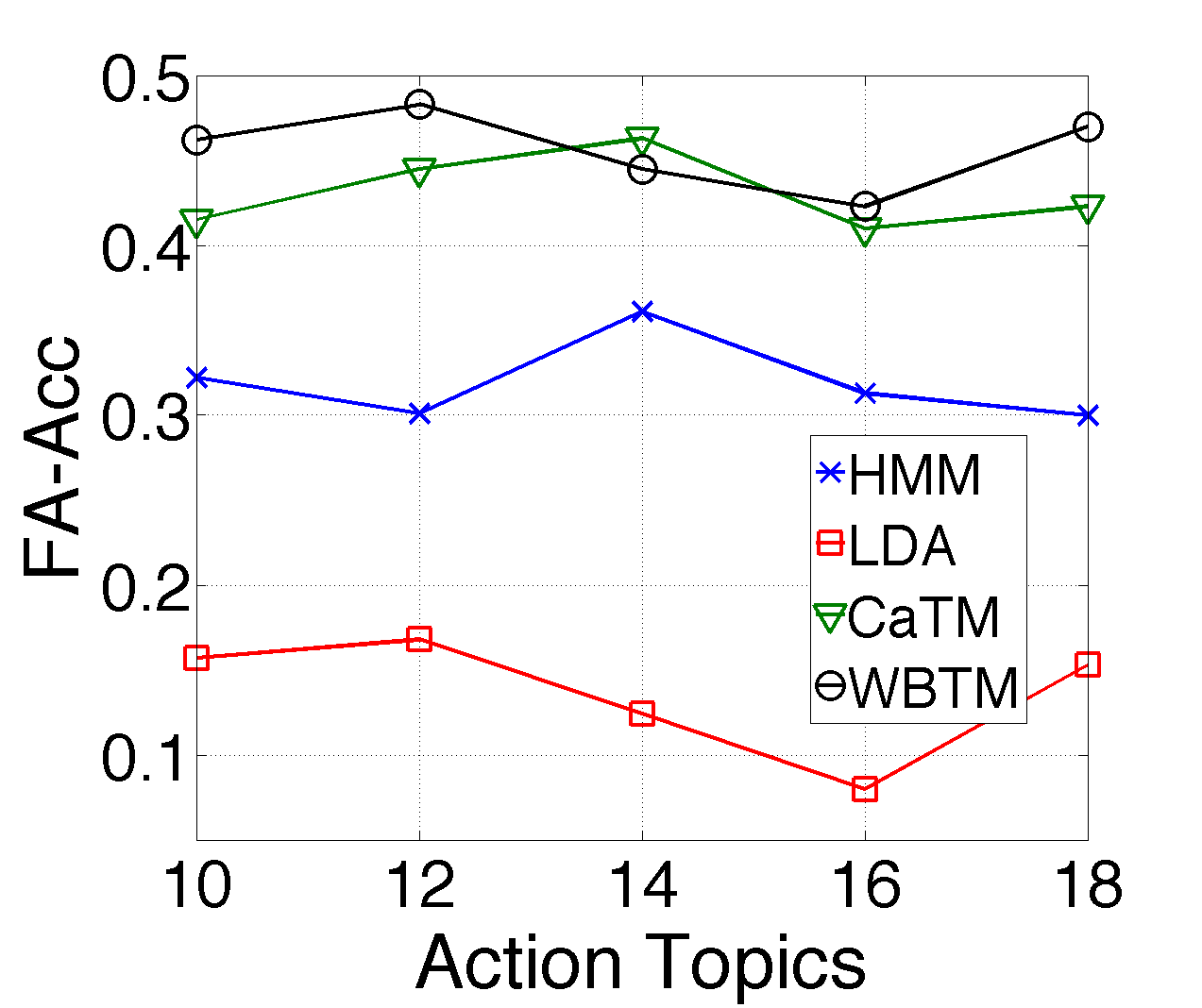}\hspace{-0.15in}
  \includegraphics[width=0.54\linewidth]{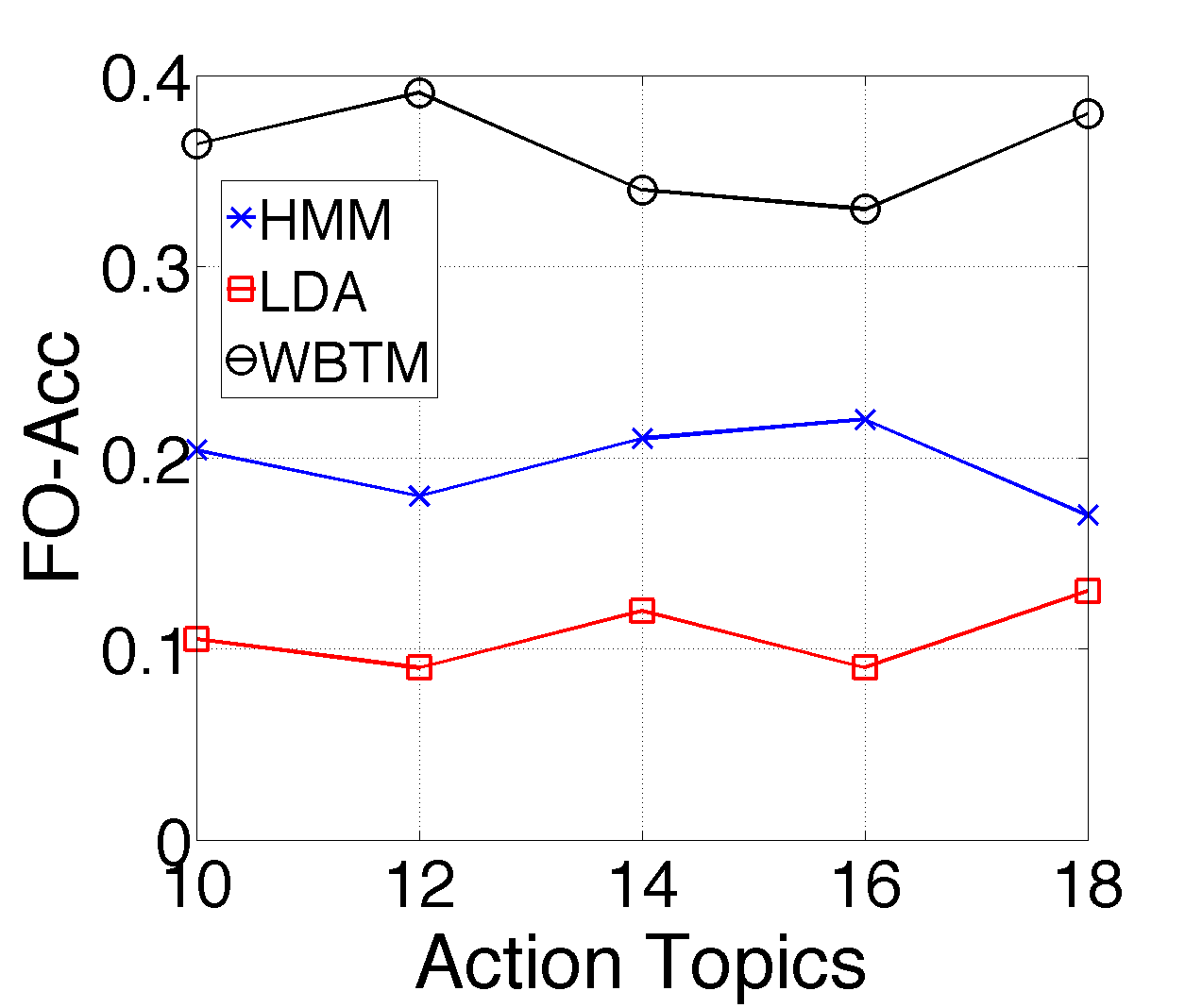}\hspace{-0.3in}
  \hspace{-0.2in}
  \vspace{-0.1in}
 \caption{Forgotten action/object detection accuracy varied with the number of action-topics in `office' dataset.} \label{fig:acp}
 \end{center}
 \vspace{-0.1in}
\end{figure}

Table~\ref{tb:re}, Fig.~\ref{fig:act} and Fig.~\ref{fig:acp} show the main results of our experiments. %We can see that our approach outperforms other baseline approaches in unsupervised action segmentation and cluster assignment, as well as forgotten action/object detection. 
We discuss our results in the light of the following questions.

{\bf How well did forgotten action/object detection perform?}
In Table~\ref{tb:re}, we can see that our model achieves a promising results for complex activities with multiple objects in variant environments in the completely unsupervised setting. Our models CaTM and WBTM show better performance than traditional uncorrelated topic model LDA, since the co-occurrence and temporal structure are well learned. They outperform HMM, since we consider both the short-range and long-range action relations while HMM only considers the local neighbouring states transitions. Our WBTM model improves the performance over CaTM on action clustering and forgotten action detection, also is able to detect the forgotten object, because action and object topics are factorized and their relations are well modeled.

{\bf How important is it to consider relations between actions and objects?}
From the results, we can see that the model which did well in forgotten action detection also performed well in detecting forgotten object. Since our model well considers the relations between the action and the object, it shows better performance in both forgotten action and forgotten object detection than HMM and LDA which models action and object independently as well as CaTM which only models the actions.

\begin{figure*}[t]
  \begin{center}
  \includegraphics[width=0.2\linewidth]{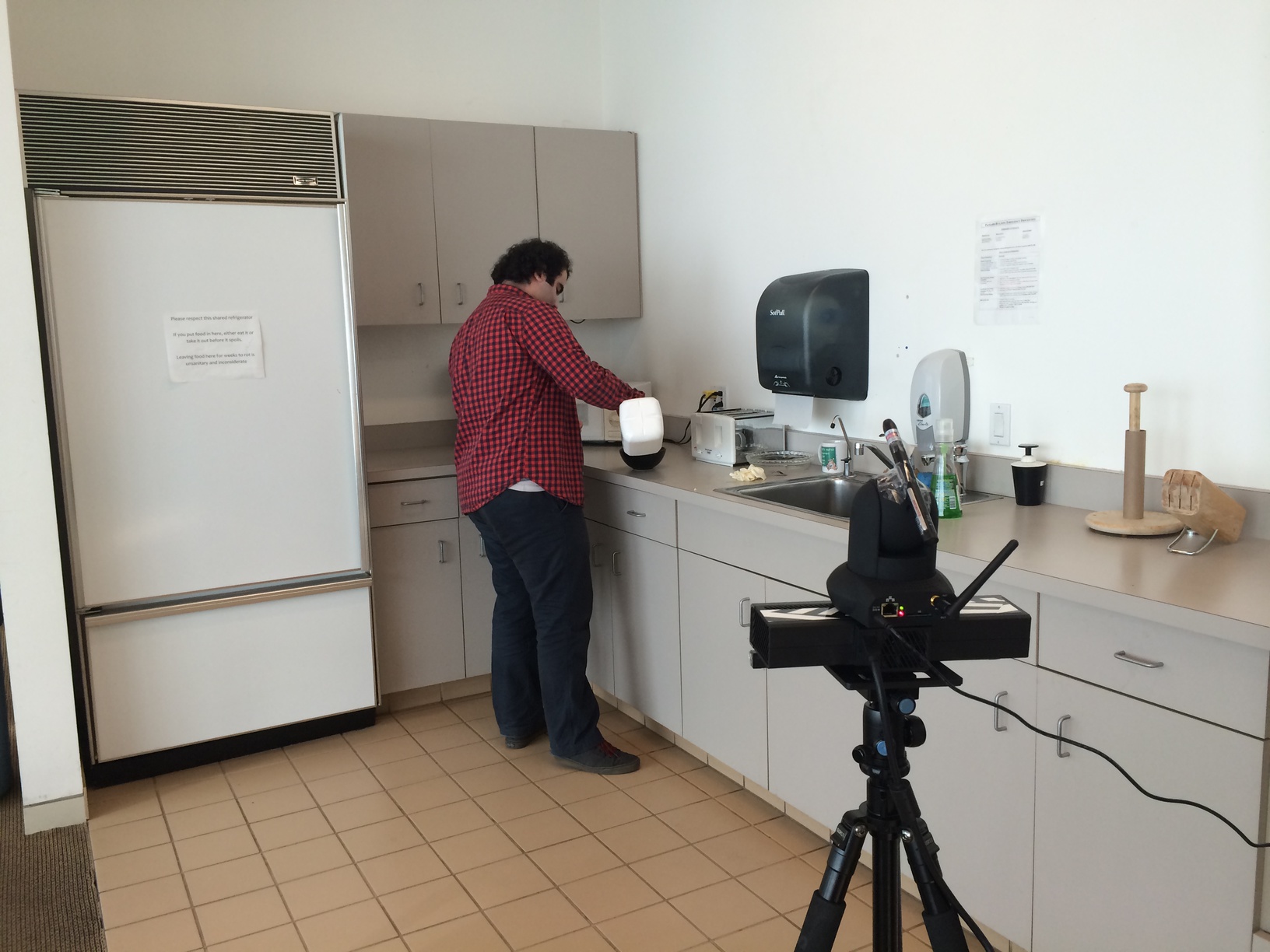}
  \includegraphics[width=0.2\linewidth]{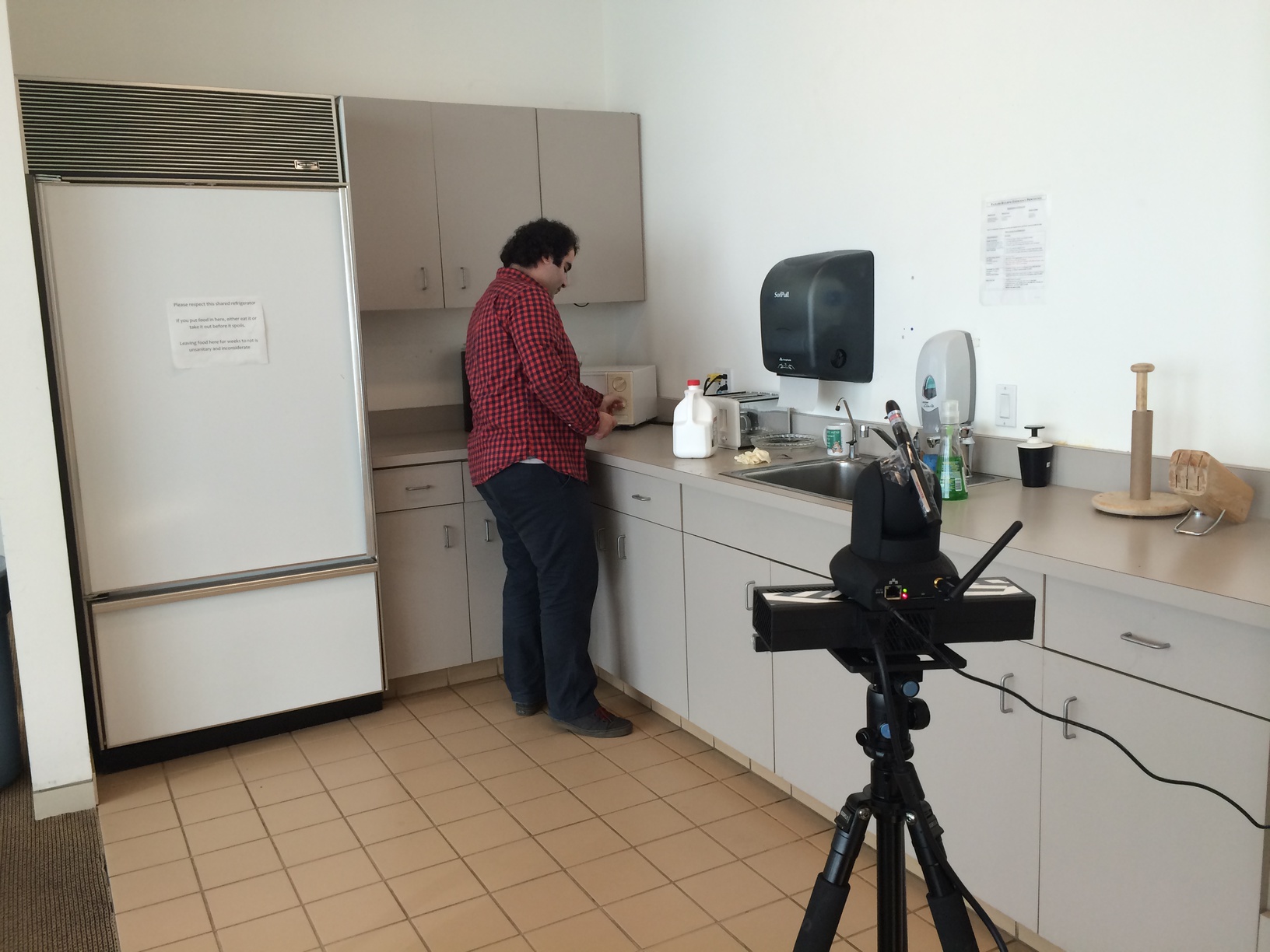}
  \includegraphics[width=0.2\linewidth]{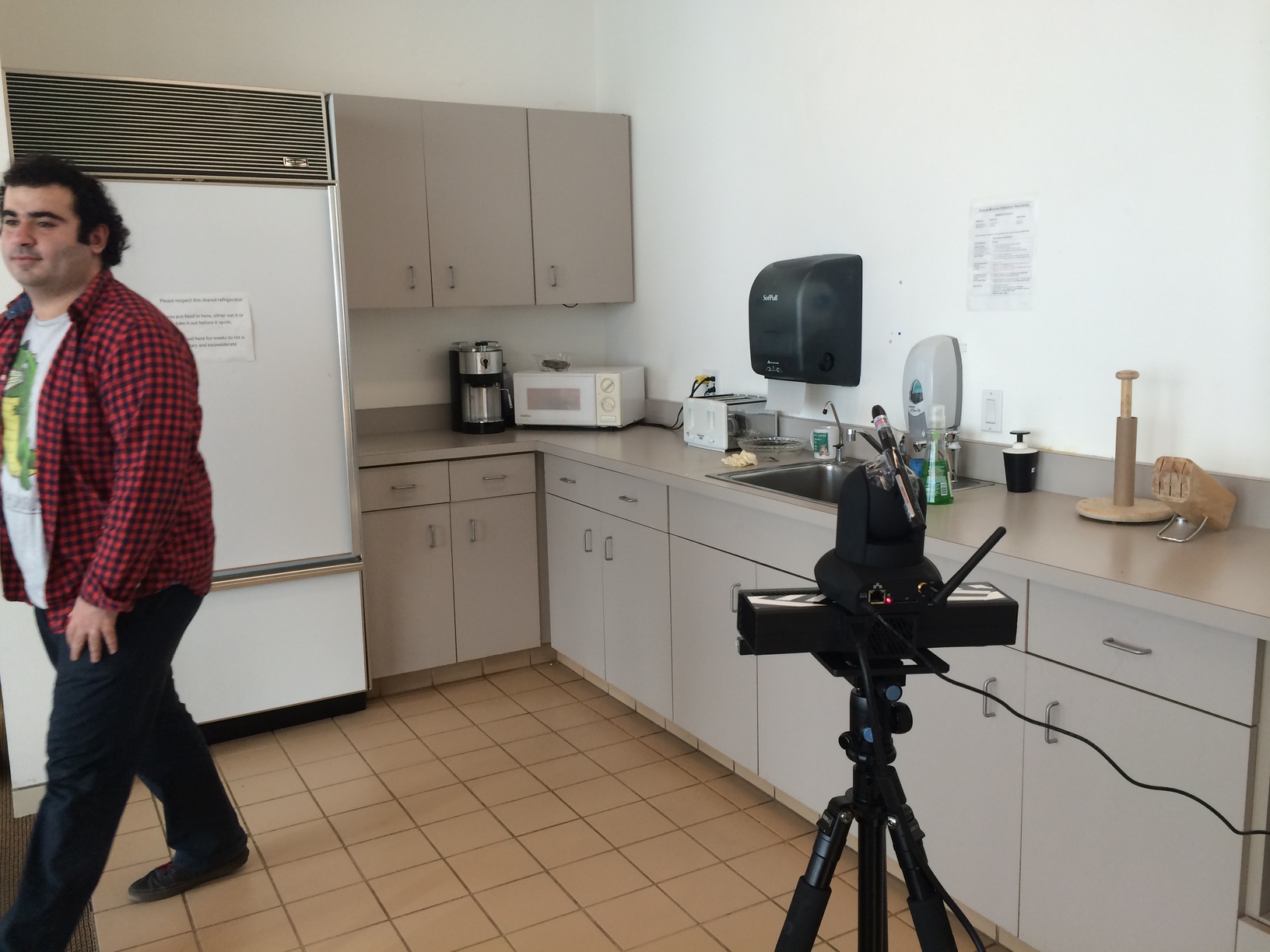}
  \includegraphics[width=0.2\linewidth]{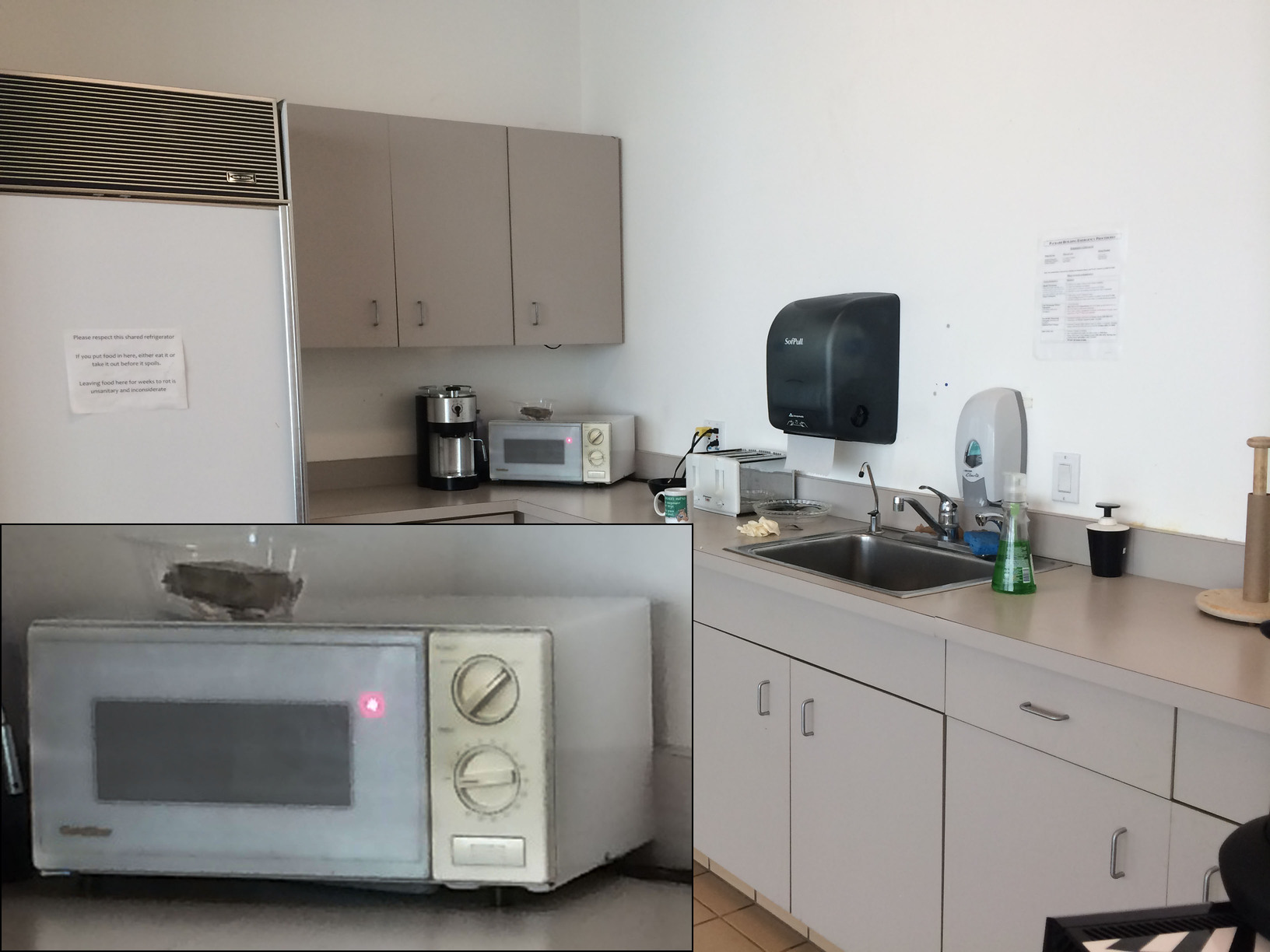}
  
 \caption{An example of the robotic experiment. The robot detects the human left the food in the microwave, then points to the microwave.} \label{fig:robotic}
 \end{center}
 \vspace{-0.3in}
\end{figure*}

%if we set the action-topics a bit more than ground-truth classes, the performance increases since a certain action might be divided into several action-topics. But the performance would saturate as action-topics increase because more variations are also introduced.

%We studied whether modeling the correlations and the temporal relations between topics was useful.
%The approaches considering the temporal relations, HMM, TM-RT, and our CaTM, outperform other approaches which assume actions are temporal independent. This demonstrates that understanding temporal structure is critical to recognizing and patching actions. The approaches, TM-RT and CaTM, which model both the short-range and the long-range relations perform better than HMM only modeling local relations. Also, the approaches considering the topic correlations CTM, CTM-AT, and our CaTM perform better than the corresponding non-correlated topic models TM, TM-AT, and TM-RT. Our CaTM, which considers both the action correlation priors and the temporal relations, shows the best performance.

%We also plot the relative time distributions learned by our model and fitted by the groundtruth labeled data in Fig.~\ref{fig:tpdf}. We can see that the distributions we learned can correctly estimate the order of the actions and the shape is mostly similar to the real distribution.

{\bf How successful was our unsupervised approach in learning meaningful action-topics?} % corresponding to semantic actions?}
From Table~\ref{tb:re} and Fig~\ref{fig:act}, we can see that the unsupervised learned action-topics can be semantic meaningful even though ground-truth semantic labels are not provided in the training. It can also be seen that, the better action segmentation and cluster assignment performance often indicates better forgotten action detection performance, since actions in the complex activity should be first well segmented and discriminated for next stage forgotten action/object detection.

{\bf How did the performance change with the number of action-topics?}
We plot the performance curves varied with the action-topic number in Fig.~\ref{fig:act} and Fig.~\ref{fig:acp}. It shows that the performance does not change much with the action-topics. This is because a certain action might be divided into several action-topics but more variations are also introduced.

\begin{table}[t]
\footnotesize
\setlength{\tabcolsep}{6pt}
\begin{center}
\caption{Robotic experiment results. The higher the better.}\label{tb:robore}
\begin{tabular*}{\linewidth}{@{\extracolsep{\fill}}c|c c c c}
\hline
&Succ-Rate(\%)&Subj-AccScore(1-5)&Subj-HelpScore(1-5)\\
\hline\hline
HMM&37.5&2.1&2.3\\
LDA&29.2&1.8&2.0\\
WBTM&\textbf{62.5}&\textbf{3.5}&\textbf{3.9}\\
\hline
\end{tabular*}
\end{center}
\vspace{-0.1in}
\end{table}

\vspace{-0.1in}
\subsection{Robotic Experiments}
\vspace{-0.05in}
In this section, we show how our Watch-Bot reminds people of the forgotten actions in the real-world scenarios.  We test each two forgotten scenarios in `office' and `kitchen' respectively (\emph{put-back-book}, \emph{turn-off-monitor}, \emph{put-milk-back-to-fridge} and \emph{fetch-food-from-microwave}). We use a subset of the dataset to train the model in each activity type separately. In each scenario, we ask $3$ subjects to perform the activity twice. Therefore, we test $24$ trials in total. We evaluate three aspects. One is objective, the success rate (Succ-Rate): the laser spot lying within the object as correct. The other two are subjective, the average Subjective Accuracy Score (Subj-AccScore): we ask the participant if he thinks the pointed object is correct; and the average Subjective Helpfulness Score (Subj-HelpScore): we ask the participant if the output of the robot is helpful. Both of them are in $1-5$ scale, the higher the better.

Table~\ref{tb:robore} gives the results of our robotic experiments. We can see that our robot can achieve over $60\%$ success rate and gives the best performance. In most cases people think our robot is able to help them understand what is forgotten. Fig.~\ref{fig:robotic} gives an example of our experiment, in which our robot observed what a human is currently doing, realized he forgot to fetch food from microwave and then correctly pointed out the microwave in the scene.

%find that approaches with better recognition mostly show better patching performance.
%\todo{From Table~\ref{tb:re}, we find that approaches with better recognition mostly show better patching performance. However, action patching can still be achieved using probabilistic inference without very accurate understanding of every action in the video. This is because patching only require missing content in the learned co-occurrence and temporal structure rather than fully correct segmentation and semantic labeling of actions.}

 %A visual example of patching is shown in Fig.~\ref{fig:vp}. It can be seen that, from the output patched segments, people can clearly know which action is forgotten.

%% file: conclusion.tex
% !TEX root = patch.tex

\vspace{-0.1in}
\section{Conclusion}\label{sec:cc}

In this paper, we enabled a Watch-Robot to automatically detect people's forgotten actions. We showed that our robot is easy to setup and our model can be trained with completely unlabeled videos without any annotations. We modeled an activity video as a sequence of action segments, which we can understand as meaningful actions.
%  have been taken.
We modeled the co-occurrence between actions and the interactive objects as well as the temporal relations between these segmented actions. Using the learned relations, we inferred the forgotten actions and localized the related objects. We showed that our approach improved the unsupervised action segmentation and cluster assignment performance, and was able to detect the forgotten action on a complex human activity \mbox{RGB-D} video dataset. We showed that our robot was able to remind people of forgotten actions in the real-world robotic experiments by pointing out the related object using the laser pointer.